\documentclass[APA,STIX2COL]{WileyNJD-v2}
\usepackage{subfig}

\articletype{Article Type}%

\received{6 March 2023}
\revised{6 June 2016}
\accepted{6 June 2016}

\begin{document}

\title{Neural operator for structural simulation and bridge health monitoring\protect}

\author[1]{Chawit Kaewnuratchadasorn $\dagger$}%
\author[2]{Jiaji Wang  $\dagger$}%
\author[1]{Chul-Woo Kim*}%

\authormark{Kaewnuratchadasorn C. \textsc{et al}}

\address[1]{\orgdiv{Department of Civil and Earth Resources Engineering}, \orgname{Kyoto University}, \orgaddress{\state{Kyoto}, \country{Japan}}}

\address[2]{\orgdiv{Department of Civil Engineering}, \orgname{The University of Hong Kong}, \orgaddress{\state{Pok Fu Lam}, \country{Hong Kong}}}

\corres{*Chul-Woo Kim. \email{kim.chulwoo.5u@kyoto-u.ac.jp }}

\presentaddress{C1-183, KyotoDaigaku-Katsura, Nishikyoku, Kyoto 615-8540, Japan}

\abstract[Summary]{Infusing deep learning with structural engineering has received widespread attention for both forward problems (structural simulation) and inverse problems (structural health monitoring). Based on Fourier Neural Operator, this study proposes VINO (Vehicle-bridge Interaction Neural Operator) to serve as the digital twin of bridge structures. VINO learns mappings between structural response fields and damage fields. In this study, VBI-FE dataset was established by running parametric finite element (FE) simulations considering a random distribution of structural initial damage field. Subsequently, VBI-EXP dataset was produced by conducting an experimental study under four damage scenarios. After VINO was pre-trained by VBI-FE and fine-tuned by VBI-EXP from the bridge at the healthy state, the model achieved the following two improvements. First, forward VINO can predict structural responses from damage field inputs more accurately than the FE model. Second, inverse VINO can determine, localize, and quantify damages in all scenarios, suggesting the practicality of data-driven approaches.}

\keywords{Fourier Neural Operator, Vehicle-Bridge Interaction, Structural Responses, Damage Fields, Structural Simulation, Structural Health Monitoring}


\maketitle

\footnotetext{$\dagger$ These authors contributed equally to this work.}

\section{Introduction}\label{sec:introduction}

Because of the combination of repeated external loads, environmental degradation, earthquakes, and other disasters, the structural performance may be decreasing. Structural Health Monitoring (SHM) has become an important discipline in civil engineering, which aims to identify anomalies and detect structural degradation to mitigate deterioration or even collapses of structures \citep{brownjohn}. In transportation systems, bridges widely experience increased usage due to loads of growing traffic, wind, temperature, earthquake, or other environmental effects. Therefore, Bridge Health Monitoring (BHM) has gained growing attention for maintenance purposes \citep{rizzo, kot}. 

Visual inspection and vibration-based BHM are two research topics in evaluating the performance of existing bridge structures. Conventional visual inspections are typically based on human visual inspections, while the computer vision-based inspection of bridges has also been developed recently. The visual inspections focus on detecting concrete cracking, concrete crushing, steel corrosion, and steel fracture of visible components. Conventional visual inspections are labor-intensive, time-consuming, traffic-interfering, and high-cost \citep{hou, flah}. In addition, the visual inspection results may be subjective to inspectors’ judgments \citep{an}. The vibration-based BHM adopts a data acquisition system to obtain vibration signals (i.e. displacement, rotation angle, acceleration, and strain) of the bridge through various sensors, including displacement sensors, gyro sensors, accelerometers, and strain gauges. Vibration-based BHM can detect the damage in bridge structures including both visible and invisible components. The advancements in data acquisition software and hardware have driven vibration-based BHM to become a promising solution for investigators, project managers, and infrastructure operators at the industrial level \citep{gharehbaghi,jahangir}. In BHM, the damage detection may be classified into four levels: Level 1 determines the existence of damage on a bridge; Level 2 localizes the damage along the span of the bridge; Level 3 quantifies the damage severity either globally or locally; and Level 4 predicts the remaining operation lifespan of the bridge. Most of the research focuses on the first three identification levels \citep{gharehbaghi,hou}.

BHM algorithms may be divided into model-driven (i.e. physics-informed) algorithms and data-driven algorithms. The BHM algorithms may also be divided into time-domain algorithms and frequency-domain algorithms. The frequency-domain algorithms generally identify the changes in modal parameters (i.e. frequency and modal shape) induced by damages and predict the damage distribution based on modal parameters. A number of research works have reported successful identifications of modal parameters through model updating or signal processing. For instance, \citet{jahangir} obtained mode shape and modal strains by Wavelet transform and classified the damaged bridge responses from the undamaged bridge responses. Other research \citep{cao,abeykoon} analyzed the damping coefficient from acceleration responses and classified undamaged and damaged structures by the differences in the coefficient. \citet{chang} identified frequencies and modal damping ratios of each mode using Multi-variate Auto-Regressive (MAR) model and detect the damages by Mahalanobis distance of collected modal parameters. However, the frequency-domain algorithms may not fully utilize the measured data in the time domain. In addition, the modal parameters may be too sensitive to environmental effects and it may be hard to infer the damage distribution based on the identification results of modal parameters \citep{yang, an, kot}.

Recent years have seen the accelerated development of Artificial Intelligence (AI) and its application to science and engineering. The machine learning and deep learning approaches also gained more attention in the structural monitoring fields. \citet{wangjiaji} compared and suggested that DeepLabv3+ with the ResNet101 backbone showed the greatest performance among all five state-of-the-art architectures and three backbones on the Crackv1 dataset in crack detection for bridge monitoring purposes. In vibration-based SHM, many works provided a justification between undamaged and damaged states of a bridge through frameworks that integrate machine learning and the critical index for damages \citep{ulrike, alireze, ana}. \citet{goi} applied principal component analysis (PCA) on a vector autoregressive (VAR) model to extract features for the proposed damage index based on the hypothesis test. \citet{bao} converted time-series responses into images and applied Deep Neural Network (DNN) to classify seven anomaly patterns (normal, missing, minor, outlier, square, trend, and drift), achieving 87 \% accuracy on the test set. \citet{luo} applied window frame on long-term time-series data and classified impulse responses by Deep Auto-Encoder (DAE), then indicated deterioration process by bridge health index. \citet{paolo} simulated spacecraft structure and added damages to generate structural responses. Then, the LSTM network was utilized to predict the damage scenario at the numerical level. \citet{avci} reviewed the work conducted in the domain, which mainly employed Artificial, Fuzzy, and Convolution Neural Networks (ANN, FNN, and CNN) with feature extraction and data processing to detect and localize damages; although most of the reviewed work reached over 90\% accuracy, all required training data from both healthy structure and damaged structure. In addition, machine learning algorithms may be divided into supervised learning and unsupervised learning. The unsupervised learning models are more convenient to apply but difficult to determine the sensitivities, limitations, and practicality. The supervised models currently encounter difficulty in developing well-established datasets with labels of damage and are mostly validated only based on numerical datasets instead of experimental datasets \citep{gomez}. The supervised learning models may need training data of damaged structures, which may be impractical for BHM application \citep{azimi, toh, malekloo, meisam}.

In response to the aforementioned challenges in BHM, this paper proposes the Vehicle-bridge Interaction Neural Operator (VINO) framework for data-driven SHM and structural simulation. VINO adopts the Fourier Neural Operator (FNO) \citep{zongyi} architecture and is trained on the Vehicle Bridge Interaction (VBI) dataset for the damage detection problem. As a new benchmark of deep learning architecture in solving partial differential equations, FNO is an encoder-decoded-based model which is able to learn function mapping \citep{zongyi, nikola}. Two VBI datasets are generated in this study, including the numerical VBI dataset based on finite element (FE) analysis (VBI-FE) and the experimental VBI dataset based on laboratory experiments (VBI-EXP) on a scaled bridge. The fine-tuning approach is used to achieve damage detection of the scaled bridge. The pre-trained model from the Vehicle-Bridge Interaction Finite Element (VBI-FE) dataset was fine-tuned only by experimental data from a healthy bridge (VBI-EXP dataset) to predict the data on the bridge at the damaged state. Therefore, the contributions are as follows:

\begin{enumerate}
    \item This study introduces the Vehicle-bridge Interaction Neural Operator (VINO), which is an end-to-end framework to detect damage directly from the structural response and predict structural response directly from damage distribution. The VINO can be more accurate and faster than the FE model in predicting structural responses.
    \item This study achieves a real-time all-in-one damage determination, localization, and quantification model by mapping between the structural damage field and the structural response field. The inverse VINO model map from the structural damage field to the structural response field. 
    \item This study predicts damages on a bridge at damaged states with only experimental data of healthy bridges using the fine-tuning method on the pre-trained model.
\end{enumerate}

In this paper, section 2 describes the methodologies of VBI, FNO, and transfer learning in the proposed framework VINO. Then, section 3 reveals the numerical (VBI-FE) and experimental (VBI-EXP) datasets for training, testing, fine-tuning, and validating VINO models. The numerical results of forward and inverse VINO for structural simulation and SHM are discussed in section 4, and the performances on experimental data are reported in section 5. Lastly, section 6 summarizes the contribution of work and provides suggested future work for more practical data-driven research in bridge and structural engineering.

\section{Methodologies}

\subsection{Background of Vehicle-Bridge Interaction}
\label{sec:VBI}

Vehicle Bridge Interaction (VBI) means the interaction between a moving vehicle and a bridge. To simulate the VBI effect, models typically consist of three main input components: bridge, vehicle, and road profile which influence the outputs differently. The outputs from the simulation are the responses of the bridge and vehicle (displacement, velocity, and acceleration) at distinct time steps \citep{clough, kimchulwoo, yang}. 

For the bridge, researchers apply different types of bridge models in an FE model for specific applications, including beam element models, shell element models, solid element models, and hybrid models. In this study, the Euler-Bernoulli beam with simple support is considered to generate the output for machine learning applications. The parameters of the input bridge include the span length, number of beam elements, bridge mass per unit length, damping coefficient, elastic modulus, and moment of inertia. For damage detection purposes, a damage field (the distribution of damage along bridge length) can be added to the parameters to obtain a simulation of the damaged bridge. 

Various complexities of vehicle models have been seen in past studies. Vehicle models range from a simple single-vehicle force which refers to a constant force moving on a bridge, to more complicated car models. Options in car models include quarter-car, half-car, and full-car, with different degrees of freedom. Therefore, the required parameters in the vehicle model include mass, the moment of inertia, the damping constant, the stiffness constant, the distance between axles, and vehicle speeds. Indeed, each component has a particular effect on bridge and vehicle responses in the simulation. 

The road profiles (often referred to as road surface roughness) affect the dynamic responses of bridges and vehicles. ISO 8680 classified road profiles into eight classes from A to H (best to poorest). The profile is derived as a representative function of road surface roughness $r(x)$ as written in Equation \ref{eq:roadprofile} \citep{iso8608}.

\begin{equation}\label{eq:roadprofile}
    r(x) = \Sigma \ d_i \left( \cos n_i x + \theta_i \right)
\end{equation}

where $x$ is the position along the bridge span; $n_i$ denotes the $i$th spatial frequency; $d_1$ and $\theta_1$ represent the roughness amplitude and phase angle, respectively. The roughness amplitude is determined for each class in Equation \ref{eq:roughness}.

\begin{equation}\label{eq:roughness}
    d = \sqrt{2G_d (n)\Delta n}
\end{equation}

where $\Delta n$ indicates the sampling interval and $G_d(n)$ is the power spectral density (PSD) function of each road class. According to ISO 8680, the PSD function is derived in Equation \ref{eq:psdfunc} as follows.

\begin{equation}\label{eq:psdfunc}
    G_d(n) = G_d(n_0)(n/n_0)^{-w}
\end{equation}

In Equation \ref{eq:psdfunc}, $G_d(n_0)$ is determined by the roughness class in ISO8680; $n$ represents the spatial frequency per meter; $w$ is 2; and $n_0$ is 0.1 cycle per meter. For instance, class A and class B roads consider $0.001\times10^{-6}$ m\textsuperscript{3} and $8\times10^{-6}$ m\textsuperscript{3} for roughness $G_d(n_0=0.1)$, respectively.

\begin{figure*}
\centerline{\includegraphics[width=0.6\linewidth]{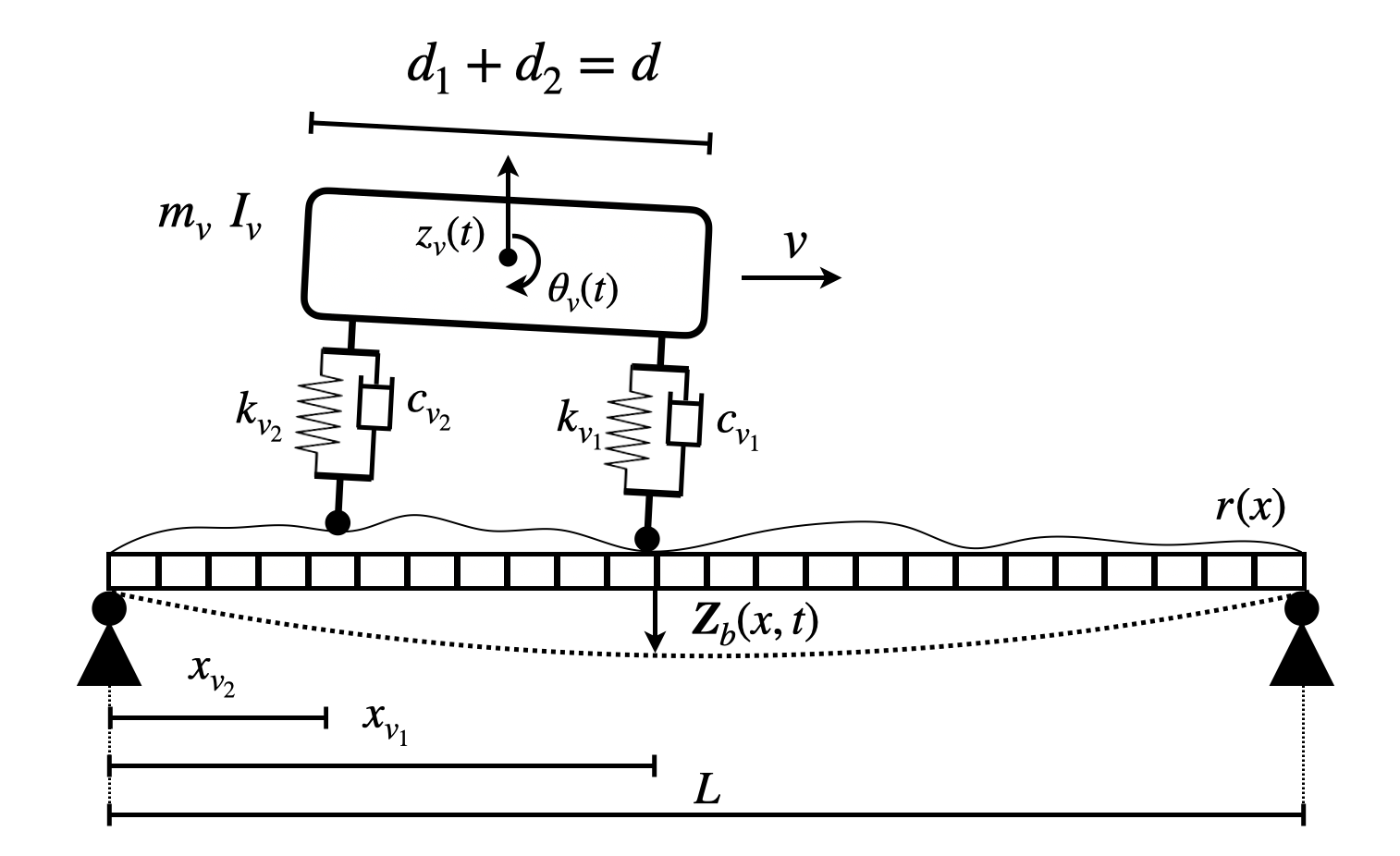}}
\caption{Vehicle-Bridge Interaction system consisting of a half-car model, element-segmented bridge, road profile}\label{fig:VBI}
\end{figure*}

Figure \ref{fig:VBI} demonstrates the interaction of vehicle and bridge. The governing equations for the 2-degree-of-freedom half-car model are shown in Equations \ref{eq:bridgeeq}-\ref{eq:veheq} to describe the system. In Figure \ref{fig:VBI}, $m_v$ and $I_v$ stand for mass and moment of inertia of the vehicle; $d_1$ and $d_2$ are the space distance between the center of mass to the first and second axle, orderly;  and $k_{v_1}$ and $k_{v_2}$ symbolizes the spring constant of the axles; $c_{v_1}$ and $c_{v_2}$ denote the damping constant of the axles; $x_{v_1}$ and $x_{v_2}$ are the locations of the axles on the bridge; $z_v(t)$ represents the displacement of the vehicle in z-axis direction; $r(x)$ specifies roughness at position $x$; and $L$ designates bridge span length.

\begin{equation}\label{eq:bridgeeq}
\begin{aligned}
0= & m_v \ddot{z}(t) \\
&+c_{v_1}\left[\dot{z}_v(t) + d_1\dot{\theta_v}(t) - \left\{ \mathbf{l_b}(x_{v_1}) \right\}^T \left\{\dot{\mathbf{Z}}_b (t) + vr'(x_{v_1}) \right\}\right]\\
&+c_{v_2}\left[\dot{z}_v(t) + d_2\dot{\theta_v}(t) - \left\{ \mathbf{l_b}(x_{v_2}) \right\}^T \left\{\dot{\mathbf{Z}}_b (t) + vr'(x_{v_2}) \right\}\right] \\
&+k_{v_2} \left[z_v(t) + d_1\theta_v(t) - \left\{ \mathbf{l_b}(x_{v_1}) \right\}^T \left\{\mathbf{Z}_b (t) + r(x_{v_1}) \right\}\right] \\
&+k_{v_2} \left[z_v(t) + d_2\theta_v(t) - \left\{ \mathbf{l_b}(x_{v_2}) \right\}^T \left\{\mathbf{Z}_b (t) + r(x_{v_2}) \right\}\right] 
\end{aligned}
\end{equation}
\begin{equation}\label{eq:veheq}
\begin{aligned}
0=&I_v\ddot{\theta}_v(t)\\
&+d_1c_{v_1}\left[\dot{x}_v(t) + d_1\dot{\theta}_v(t) - \left\{ \mathbf{l_b}(x_{v_1})\right\}^T \left\{ \dot{\mathbf{Z}}_b(t) \right\} + vr'(x_{v_1}) \right]\\
&-d_2c_{v_2}\left[\dot{x}_v(t) + d_2\dot{\theta}_v(t) - \left\{ \mathbf{l_b}(x_{v_2})\right\}^T \left\{ \dot{\mathbf{Z}}_b(t) \right\} + vr'(x_{v_2}) \right] \\
&+d_1k_{v_1}\left[x_v(t) + d_1\theta_v(t) - \left\{ \mathbf{l_b}(x_{v_1})\right\}^T \left\{ \mathbf{Z}_b(t) \right\} + r(x_{v_1}) \right]\\
&-d_2k_{v_2}\left[x_v(t) + d_1\theta_v(t) - \left\{ \mathbf{l_b}(x_{v_2})\right\}^T \left\{ \mathbf{Z}_b(t) \right\} + r(x_{v_2}) \right]
\end{aligned}
\end{equation}

where ${\mathbf{Z}_b(t)} = \left\{Z_{b_1}(t), Z_{b_2}(t), \dots, Z_{b_{n_b}}(t) \right\}$ is the vector of displacement each node from $1$ to $n_b$ on the bridge system. $\left\{ \mathbf{l_b}(x_{v_i})\right\}$ designates the vector that contains polynomial interpolation functions for the displacement of the bridge system observed at the contact point of the $i$th axle. The over-dot and prime symbolize the derivative with respect to time and space, respectively. 

The dynamic equation of motion of the bridge can be expressed in Equation \ref{eq:dynamiceq} where mass, stiffness, and damping matrices $\left[ \mathbf{M}_b \right]_{2n_b\times 2n_b}$, $\left[ \mathbf{K}_b \right]_{2n_b\times 2n_b}$, $\left[ \mathbf{C}_b \right]_{2n_b\times 2n_b}$ can be constructed from the bridge parameters. 

\begin{equation}\label{eq:dynamiceq}
    \begin{aligned}
    \left[ \mathbf{M}_b\right]\left\{ \ddot{\mathbf{Z}}_b(t)\right\} + \left[ \mathbf{C}_b\right]\left\{ \dot{\mathbf{Z}}_b(t)\right\} + \left[ \mathbf{K}_b\right]\left\{ \mathbf{Z}_b(t)\right\}\\
    \qquad + \left\{ \mathbf{l}_b\left(x_{v_1}\right)\right\}R_1(t) + \left\{ \mathbf{l}_b\left(x_{v_2}\right)\right\}R_2(t)& = \mathbf{0}
    \end{aligned}
\end{equation}

where $R_1(t)$ and $R_2(t)$ are the contact forces at axle positions $x_{v_1}$ and $x_{v_2}$. The contact forces at each time step can be calculated as follows. 

\begin{equation}\label{eq:interact-1}
    \begin{aligned}
    R_1(t) =& - c_{v_1} \left[ \dot{z}_v(t) - \left\{ \mathbf{l_b}(x_{v_1})\right\}^T\left\{\dot{\mathbf{Z}}_b(t)\right\} + vr'(x_{v_1})\right]\\
    &-k_{v_1} \left[ z_v(t) - \left\{ \mathbf{l_b}(x_{v_1})\right\}^T\left\{\dot{\mathbf{Z}}_b(t)\right\} + r(x_{v_1})\right] \\
    &+ \left(\frac{d_2}{d}\right)m_vg
    \end{aligned}
    \end{equation}
\begin{equation}\label{eq:interact-2}
    \begin{aligned}
    R_2(t) =& - c_{v_2} \left[ \dot{z}_v(t) - \left\{ \mathbf{l_b}(x_{v_2})\right\}^T\left\{\dot{\mathbf{Z}}_b(t)\right\} + vr'(x_{v_2})\right]\\
    &-k_{v_2} \left[ z_v(t) - \left\{ \mathbf{l_b}(x_{v_2})\right\}^T\left\{\dot{\mathbf{Z}}_b(t)\right\}+ r(x_{v_2})\right]\\
    &+ \left(\frac{d_1}{d}\right)m_vg
    \end{aligned}
\end{equation}

The combination of Equation \ref{eq:bridgeeq} to Equation \ref{eq:interact-2} becomes a dynamic coupling equation that represents the whole vehicle-bridge system in Figure \ref{fig:VBI}. The equation can be solved by Newmark-$\beta$ method for implicit time integration. The method solves the displacement, velocity, and acceleration at $t+\Delta t$ from the current stage $t$ and known information at $t+\Delta t$ by Equation \ref{eq:solveacc} to Equation \ref{eq:solvedisp} \citep{newmark}.

\begin{equation}\label{eq:solveacc}
    \begin{aligned}
    \left[ \mathbf{M} +\Delta t \gamma\Delta \mathbf{C}+\Delta t^2 \beta \mathbf{K}  \right]\left[\ddot{\mathbf{Z}}(t+\Delta t)\right] = \qquad \qquad \qquad \\
    \left\{
    \begin{matrix}
    f(t+\Delta t)
    - \mathbf{C}\left[ \dot{\mathbf{Z}}(t) + \Delta t (1- \gamma) \ddot{\mathbf{Z}}(t) \right]\\
     -\mathbf{K}\left[ \mathbf{Z}(t) + \Delta t \dot{\mathbf{Z}}(t) + \Delta t^2 \left( \frac{1}{2} -  \beta\right) \ddot{\mathbf{Z}}(t)\right]
    \end{matrix}
    \right\}
    \end{aligned}
\end{equation}

\begin{equation}\label{eq:solvevec}
\dot{\mathbf{Z}}(t+\Delta t)=\dot{\mathbf{Z}}(t) +\Delta \left(1 - \gamma\right)\ddot{\mathbf{Z}}(t) + \Delta t \gamma \ddot{\mathbf{Z}}(t +\Delta t)
\end{equation}

\begin{equation}\label{eq:solvedisp}
\begin{aligned}
\mathbf{Z}(t+\Delta t)&=\mathbf{Z}(t) +\Delta \dot{\mathbf{Z}}(t) + \\
&\Delta t^2 \left( \frac{1}{2} -\beta \right)\ddot{\mathbf{Z}}(t) +\Delta t^2 \beta \ddot{\mathbf{Z}}(t+\Delta t)
\end{aligned}
\end{equation}

Herein, $\mathbf{Z}$ is a displacement matrix that contains bridge ($\mathbf{Z}_b$) and vehicle ($z_v$) responses; $\mathbf{M}$, $\mathbf{C}$, and $\mathbf{K}$ are the systematic mass, damping, and stiffness matrices of the bridge and vehicle; $\gamma$ and $\beta$ are constant parameters of Newmark-$\beta$ method.

\subsection{Fourier Neural Operator}
\label{sec:FNO}

\citet{tianping} first proved the Universal Representation Theory of neural operators in 1995, that neural networks could be trained to fit arbitrary operators. Recently, neural operators have been rapidly developed to learn the mapping from the input field to the output field of partial differential equations \citep{nikola}. The architectures of the neural operator enhance the ability to map between functions with infinite-dimensional space \citep{zongyi,nikola}. Fourier Neural Operator (FNO) is a type of neural operator architecture that utilizes Fourier Transform in the layers. FNO was reported to achieve the best performance among existing neural operators for solving complex partial differential equations in 2021 as per \citet{zongyi}. In this section, the framework and transfer learning for Fourier Neural Operator are explained to be applied in the forward and inverse problems in structural health monitoring.

\begin{figure*}[t!]
\centering
  \includegraphics[width=0.9\linewidth]{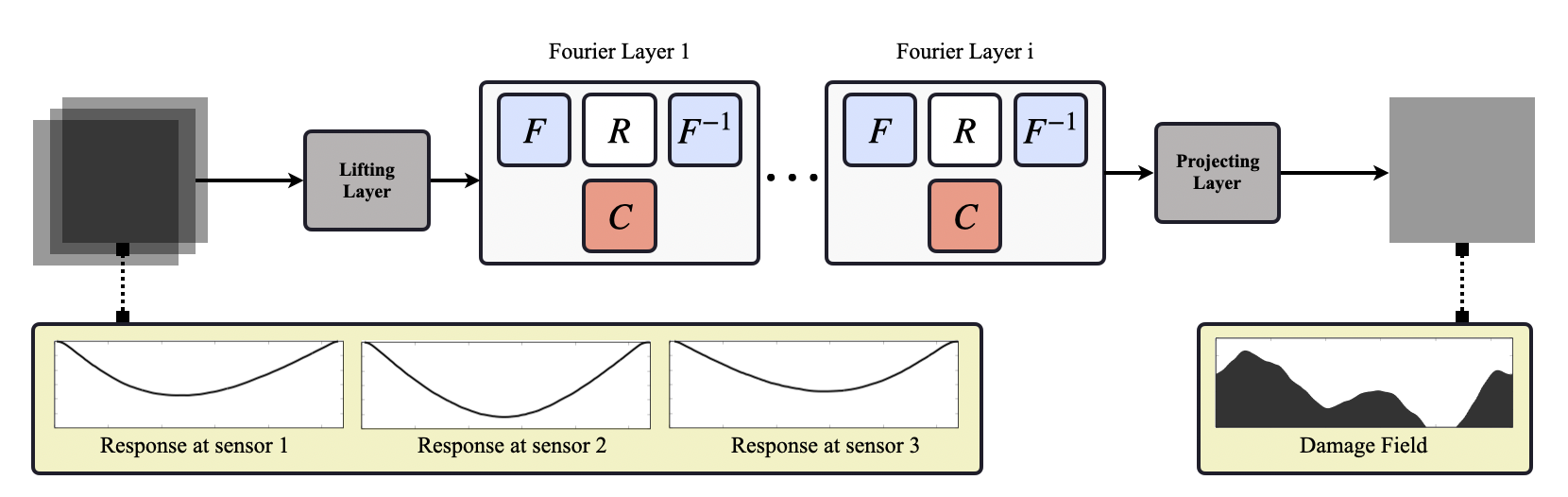}
  \caption{Fourier Neural Operator Architecture in inverse problems which composes of three main parts: lifting, Fourier, and projecting layers. In a Fourier layer, the Fourier transform ($F$), linear transform with lower-Fourier-mode filtering ($R$), inverse Fourier transform ($F^{-1}$), and convolution transform ($C$) are applied \citep{zongyi}. Herein for inverse problems, the inputs are structural responses, and the output is the damage field.}
  \label{fig:FNO}
\end{figure*}

FNO architecture comprises three main components, which are lifting, iterative kernel operator, and projecting, as demonstrated in Figure \ref{fig:FNO}. In the lifting section, the inputs of the model are lifted into the dimension of the iterative kernel operator through a linear layer. After the dimensional trajectory shift, the iterative kernel operator is a branch of connected Fourier-layer blocks. The input in each block passes along two path functions. In the first path, the input undergoes a Fourier Transform. Then, the linear layer filters out higher Fourier mode before the inverse Fourier transform operates the data to another function. In the second path, the input passes through a convolution layer. Subsequently, two paths merge to create layer output and become an input of the next layer. The last layer connects to the third part of the components — projection. The data is projected to the same dimension as the input to the model through another linear layer at the end of the architecture. Therefore, the inputs and outputs will always need to be the same dimensions.

In an iterative kernel operator, the Integral Operator (Equation \ref{eq:NO}) in Neural Operator was transformed into Fourier Operator (Equation \ref{eq:FNO}) as described in the Fourier Neural Operator architecture based on the Convolution theorem (Equation \ref{eq:convolution}).

\begin{equation}\label{eq:NO}
\left( K_t(v_t)\right)(x) := \int_D \kappa\left(x,y\right) v_t(y) dy
\end{equation}
\begin{equation}\label{eq:FNO}
\int_D \kappa(x-y)\  v_t (y) \ dy  = F^{-1}  \left( F\left( \kappa(x-y)\right) * F\left(v_t(y)\right)\right)
\end{equation}
\begin{equation}\label{eq:convolution}
\left(K(\phi)v_t\right)(x) := F^{-1}\left(R_{\phi}\cdot \left(F_{v_t} \right) \right)(x)
\end{equation}

where $R_\phi$ is the Fourier transform of a kernel function with periodic variation $\kappa$; $F$ and $F^{-1}$ are defined as Fourier and inverse Fourier transform, written in Equation \ref{eq:fourier} and Equation \ref{eq:inversefourier}.

\begin{equation}\label{eq:fourier}
    \left(Ff\right)_j(k) = \int_D f_j(x)e^{-2i\pi \langle x,k\rangle}dx
\end{equation}
\begin{equation}\label{eq:inversefourier}
    \left(F^{-1}f\right)_j(x) = \int_D f_j(k)e^{2i\pi \langle x,k\rangle}dk
\end{equation}

\subsection{Transfer Learning}
Transfer Learning refers to a technique that utilizes the weights of a neural network or neural operator learned from an existing larger dataset to a new unseen dataset or to a similar problem. The approach not only reduced the training time on the new dataset but benefited when the new dataset was insufficiently large. Recent years have seen an increase in the application of transfer learning for deep learning models in many fields. Fine-tuning is one of the transfer learning methods which trains pre-trained models on a new dataset. Since the pre-trained models usually consist of many layers, fine-tuning can be conducted in only some layers in the models; and typically, a few last layers are trained while the rest of the model is frozen \citep{chen,guillermo,chamangard}. 

In civil engineering, a computer vision-oriented pre-trained model named ResNet34 was fine-tuned on Complex Frequency Domain Assurance Criterion (CFDAC) matrix to detect alteration in stiffness. The CNN-based models such as VGG, ResNet, and AlexNet were mostly utilized in vision-based crack detection tasks as the models were pre-trained on images \citep{wangjiaji}. However, \citet{chamangard} also applied CNN-based acceleration responses with transfer learning to obtain the high accurate damage detection on insufficient data. Although the dataset was not big, the pre-trained models were also fine-tuned with damage data on the Tianjin Yonghe bridge at the detection level.

In VINO developed in this study, only the projection layers (i.e. the last two layers) of the pre-trained FNO model were fine-tuned based on an experimental dataset of the healthy bridge. The weights in the lifting and iterative kernel operator components were not fine-tuned. The fine-tuning dataset was only the responses from the health bridge obtained from experimental data. The fine-tuning approach is feasible for real-world structures because field vibration tests on the new bridge can be conducted to serve as a dataset for fine-tuning, while future vibration response of structures can be fed to FNO to detect damage distribution.

\subsection{The Proposed Framework}
\begin{figure*}[t]
\centering
\fbox{
  \includegraphics[width=\linewidth]{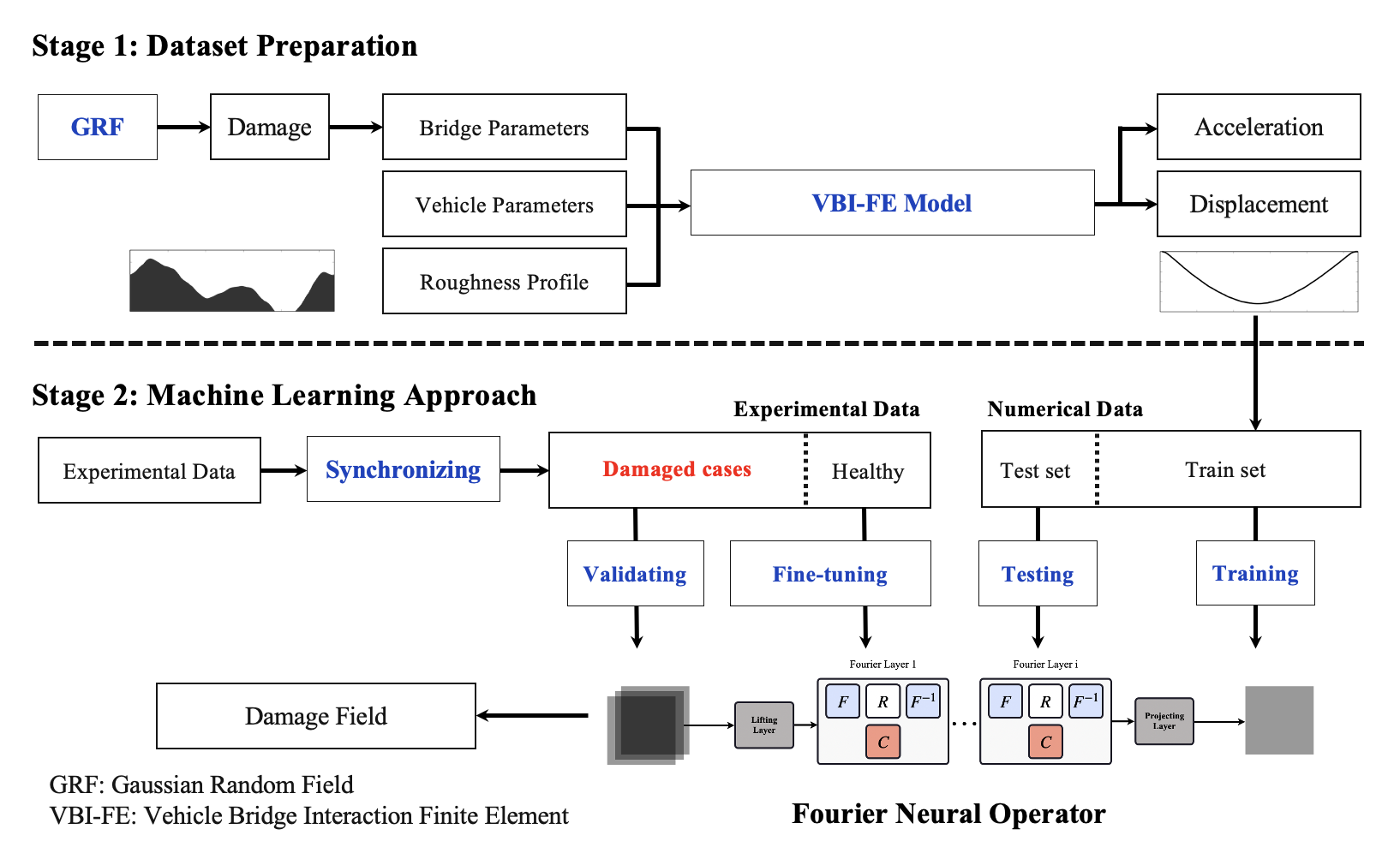}}
  \caption{The overall proposed data-driven framework and procedures}
  \label{fig:framework}
\end{figure*}

Figure \ref{fig:framework} illustrates the proposed framework to use the vehicle-bridge interaction information to pre-train and fine-tune the Fourier Neural Operator. The framework consists of two main stages, which are VBI-FE dataset preparation (stage 1) and the machine learning approach (stage 2). As previously mentioned, stage 1 adopted the vehicle-bridge interaction finite element model to generate a dataset of damage fields and structural response fields. Stage 2 focused on the model training, testing, fine-tuning, and validating where both VBI-FE and VBI-EXP datasets are used to train, verify, fine-tune, and validate the model. It is important to emphasize that in the framework, only the responses from the bridge at the healthy state were used to fine-tune in order to predict damages on the bridge at the damaged state in the same system of vehicle-bridge interaction. Additionally, this framework can serve both digital twins, depending on the inputs and outputs of the Fourier Neural Operator model. In stage 2 of Figure \ref{fig:framework}, the inputs are the structural responses while the outputs are the damage field, which can be referred to as the inverse problem for BHM.

\section{Dataset Acquisition}

\subsection{Numerical Simulation Setup for VBI-FE dataset}
\label{sec:NumSetup}

This section explains the setup of the simulation for training the VINO model. Firstly, this section will explain the model parameters in numerical simulation based on the governing equations of VBI. Then, the generation of the dataset for model training will be described by the variation of damage input to the numerical model.

\subsubsection{The numerical bridge model}
The bridge parameters came from measurements and calculations on the laboratory bridge. The Rayleigh Damping ($c$) is written in the form of fractions of mass ($M$) and stiffness ($K$) with two coefficients as shown in Equation \ref{eq:rayleigh}. The mass coefficient ($\alpha_{dM}$) and the stiffness coefficient ($\beta_{dK}$) are formulated as Equation \ref{eq:rayleigh-coef} \citep{geraschenko}. 

\begin{equation}\label{eq:rayleigh}
    c = \alpha_{dM}M + \beta_{dK} K
\end{equation}

\begin{equation}\label{eq:rayleigh-coef}
    \alpha_{dM} = 4\pi f_1f_2 \frac{\zeta_1 f_2 - \zeta_2 f_1}{f_2^2 -f_1^2} \qquad \beta_{dK} = \frac{\zeta_2 f_2 -\zeta_1f_f}{\pi \left(f_2^2 - f_1^2\right)}
\end{equation}

where $f_1$ and $f_2$ are the frequency of the first and second modes while $\zeta_1$ and $\zeta_2$ are the damping ratios which is 0.007 according to the in-house experiment described in section \ref{sec:ExpSetup}. 

\subsubsection{The numerical vehicle model}
The half-car model is adopted to simulate the vehicle-bridge interaction system. Table \ref{tab:vehicle} summarized the parameters of the vehicle measured in the in-house experiment. The speed of the vehicle on the laboratory bridge ($v=0.55 \ m/s$) is equivalently converted to the vehicle speed on the real bridge ($v=10 \ km/hr$), using the speed parameter shown in Equation \ref{eq:freqscal}.

\begin{equation}\label{eq:freqscal}
    \gamma = \frac{v}{2\cdot f_1 \cdot L}
\end{equation}

where $\gamma$ denotes the speed parameter; $v$ refers to the speed of the vehicle on the bridge with the first mode frequency ($f_1$) and length ($L$). 

Moreover, suspension damping was calculated based on the mass and stiffness of the vehicle.

\begin{equation}\label{eq:damping}
    c = c_c\times \zeta_v
\end{equation}

where $\zeta_v$ is the damping ratio obtained from previous tests of the same bridge \citep{hanzhouran}. $c_c$ indicates the critical damping which equals to $2\sqrt{m_a k_a}$ . In the calculation, the mass of each axle ($m_a$) was assumed to be half of the total mass of the vehicle ($m_v$). Thus, the damping of two axles was 45.28 $Ns/m$.

\begin{table*}[ht!]
\centering
    \captionof{table}{Bridge Parameters} \label{tab:bridge} 
    \begin{tabular*}{500pt}{@{\extracolsep\fill}lll}
        \hline
        \textbf{Parameters} & \textbf{values} & \textbf{units}\\ [1ex] 
        \hline
        Length & 5.4 & $m$  \\ 
        Mass per unit length & 53.47 & $kg/m$  \\
        Young's modulus & $2.1\times10^{11}$ & $N/m^2$    \\
        Moment of Inertia & $5.49\times10^{-7}$ & $m^4$ \\
        Frequencies ($1^{st}$ and $2^{nd}$ modes) & 3.64, 14.56 & $Hz$ \\
        Rayleigh Damping Coefficients ($\alpha_{dM}, \beta_{dK}$) & $0.2562, 1.22\times10^{-4}$ &  \\[1ex] 
        \hline
    \end{tabular*}
\end{table*}

\begin{table*}[ht!]
\centering
    \captionof{table}{Vehicle Parameters} \label{tab:vehicle} 
    \begin{tabular*}{500pt}{@{\extracolsep\fill}lll}
        \hline
        Parameters & values & units \\ [1ex] 
        \hline
        Speed & 1.35 & $m/s$  \\ 
        Space between axles & 0.3 & $m$  \\
        Sprung mass & $15.38$ & $kg$    \\
        Suspension stiffness (Axle 1, Axle 2) & 1666, 1666 & $N/m$ \\
        Suspension damping (Axle 1, Axle 2) & 45.28, 45.28 & $Ns/m$\\ 
        \hline
    \end{tabular*}
\end{table*}

\subsubsection{The road profile}
\begin{figure*}[t!]
\centering
  \includegraphics[width=0.6\linewidth]{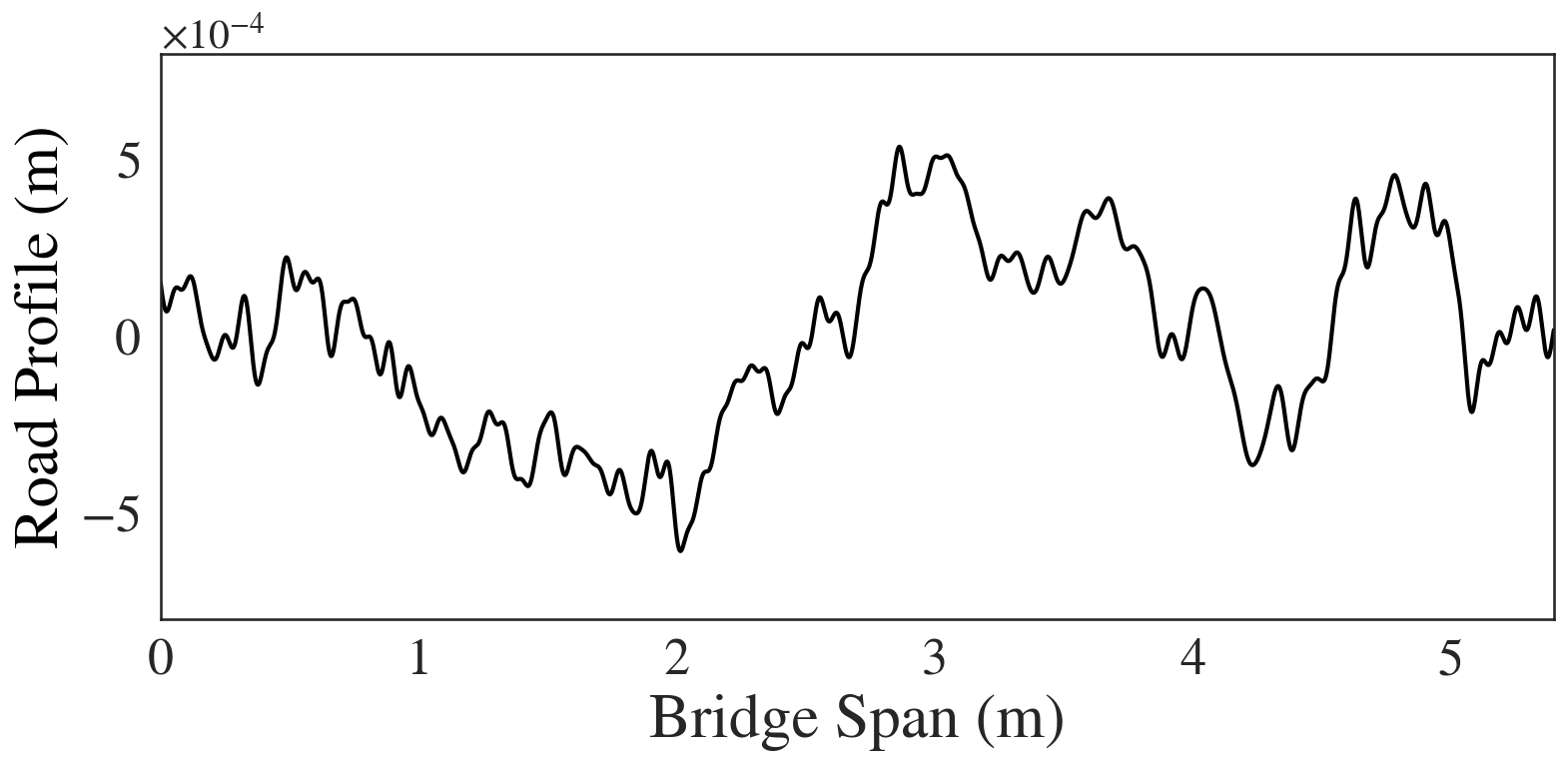}
  \caption{Road Profile on laboratory bridge}
  \label{fig:rx}
\end{figure*}

The road irregularity profile was obtained from measurements on the experimental bridge, plotted in Figure \ref{fig:rx}. 

\subsubsection{The numerical VBI-FE dataset}
The dataset stands as the most influential component in the machine learning model. In damage detection, datasets must be considered and prepared as carefully as possible regarding accuracy and practicality. Since in machine learning-application research, the models are trained on the training set and tested on the testing set, it is always possible that a training set does not provide the coverage of all possible cases, resulting in an impractical model. 

In VINO, FE simulation of a VBI system is first adopted to generate a numerical dataset VBI-FE based on the random damage possibility of every element on the bridge. The Gaussian random field is utilized to generate continuous random damage fields as written in Equation \ref{eq:grf}.

\begin{equation}\label{eq:grf}
    \begin{aligned}
    P\left(y_1, \dots, y_n\right) dy_1\dots dy_n  = \frac{1}{2\pi (\det \mathbf{M} )^{1/2}} \\
    \times \exp \left\{ -\left(y_1, \dots, y_n\right) \mathbf{M}^{-1} \left(
    \begin{matrix}
    y_1\\ \cdots\\ y_n
    \end{matrix}
    \right)/2 \right\} dy_1\dots dy_n
    \end{aligned}
\end{equation}

where $y_n$ are data points and $\mathbf{M}$ is the correlation matrix. The dataset was prepared from simulation runs on the computer spec Intel(R) Core(TM) i9-10850K CPU @3.6GHz with NVIDIA GeForce RTX 3090 at the laboratory within 24 hours with a single CPU thread. In total, the training and testing sets of VBI-FE include 1,200 independent FE simulations. This study obtained the acceleration of all 514 nodes of bridge elements and the vehicle for 844-time steps in the VBI-FE dataset. Then the dataset was divided into 1000 and 200 simulations of responses for training and testing the FNO, respectively.

\subsection{Experimental setup for VBI-EXP dataset}
\label{sec:ExpSetup}
The actual experiments were conducted in the laboratory at Kyoto University to collect real-world data in order to provide validation and justification to the Fourier Neural Operator models in both forward and inverse problems. The model bridge and model vehicle are measured and calculated for physical properties as the same values in Tables \ref{tab:bridge} and \ref{tab:vehicle}.

\begin{figure*}[t!] 
\centering
  \subfloat[Cross-sectional design of the laboratory bridge \citep{yokoyama}\label{fig:expsetup-design}]{\includegraphics[width=0.65\linewidth]{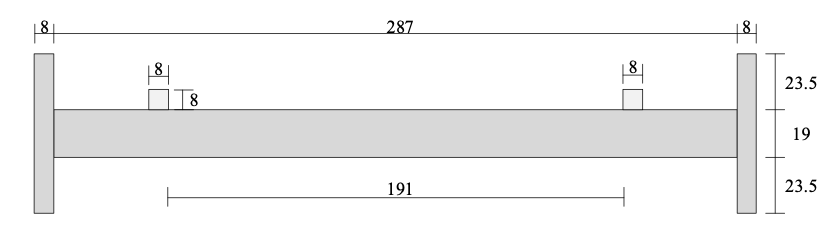}}
  \qquad 
  \subfloat[Schematic design of the laboratory bridge \citep{yokoyama} \label{fig:expsetup-design2}]{\includegraphics[width=0.9\linewidth]{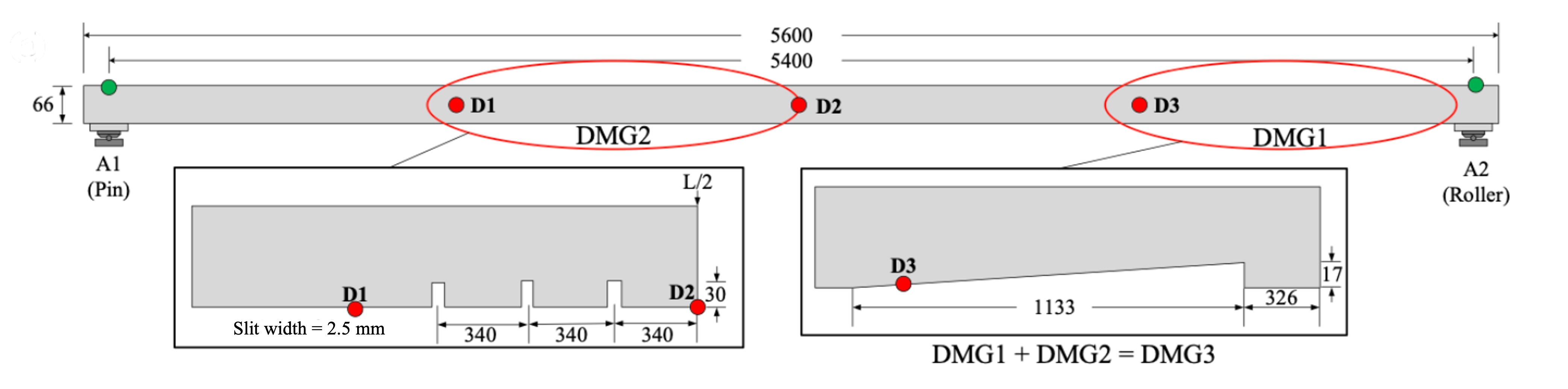}}
  \qquad
  \subfloat[Laboratory bridge \label{fig:expsetup-bridge} ]{\includegraphics[width=\linewidth]{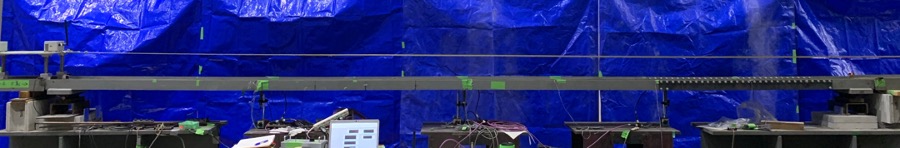}}
  \qquad
  \subfloat[The reinforcement at DMG1\label{fig:expsetup-dmg1}]{\includegraphics[width=0.45\linewidth]{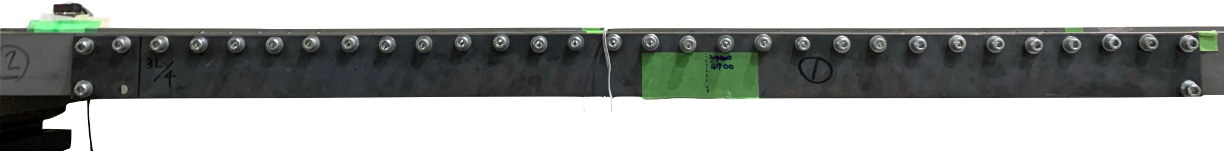}}
  \hfill
  \subfloat[The reinforcement at DMG2\label{fig:expsetup-dmg2}]{\includegraphics[width=0.45\linewidth]{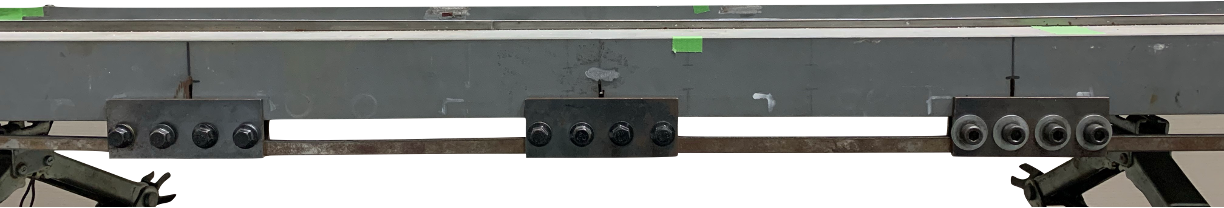}}
  \caption{Experimental setups}
  \label{fig:expsetup} 
\end{figure*}

As schematically illustrated in Figure \ref{fig:expsetup}, the model bridge has an I-section, and the weak axis is loaded in the test. The total span length is 5.4 meters. The boundary conditions of the bridge are the pin and roller supports on two ends. DMG1 (on the right) and DMG2 (on the left) were depicted in the schematic view in Figures \ref{fig:expsetup-design} and \ref{fig:expsetup-design2} \citep{yokoyama}. In this paper, the detachable reinforcement for DMG1 in Figure  \ref{fig:expsetup-dmg1} and DMG2 in Figure \ref{fig:expsetup-dmg2} were considered as no damage, where removal of the reinforcement is damage cases. As previously mentioned, four damage scenarios (INT, DMG1, DMG2, and DMG3) were assumed and corresponded to intact, damage 1, damage 2, and damage 3 conditions, whereas DMG3 is the combination of the existence of DMG1 and DMG2.

The dynamic response data of the bridge were collected from sensors placed on and under the bridge. The actual setup is shown in Figure \ref{fig:expsetup-bridge} where three displacement transducers (CDP-50mm) and three wired accelerometers (M-A552AC10) were installed at the quarter, mid, and three-quarter spans along the bridge. Two optical sensors (NPN PZ-G52N) were located at two ends in order to track the entry and exit of the vehicle in the system. The details of the vehicle were summarized in Table \ref{tab:vehicle}. Two wireless accelerometers were also deployed on top of the vehicle to monitor the acceleration of the vehicle during the experiment. All sensors were connected to the central control panel, namely DC-7204 Dynamic Strain Recorder measurement software, and synchronized to collect data from sensors and transducers. 

\section{Performance of VINO on VBI-FE Numerical Dataset}
\label{sec:NumVal}

This section provides the numerical verification of VINO for forward and inverse problems. As shown in the framework of Figure \ref{fig:framework}, the numerical dataset was divided into the training set and testing set. In this section, the performance of VINO on the testing set is reported to discuss the efficiency of the VINO. For the structural simulation problem, the forward VINO aims to generate structural responses from the input of a particular damage distribution curve, which is similar to FE simulation. For the structural health monitoring problem, the inverse VINO is validated by the accuracy in predicting damage fields from the structural responses at several sensor locations.

\subsection{Forward VINO for Structural Simulation}

Structural simulation is considered a forward problem due to the fact that it solves differential equations to obtain the responses of bridge and vehicle. The conventional model applies the FE method, as explained in section \ref{sec:VBI}, and the proposed VINO model is a data-driven architecture that learns from the VBI-FE dataset.

\begin{figure*}[ht!] 
\centering
  \subfloat[Damage field as an input to forward VINO\label{fig:numfwd-a} ]{\includegraphics[width=0.4\linewidth]{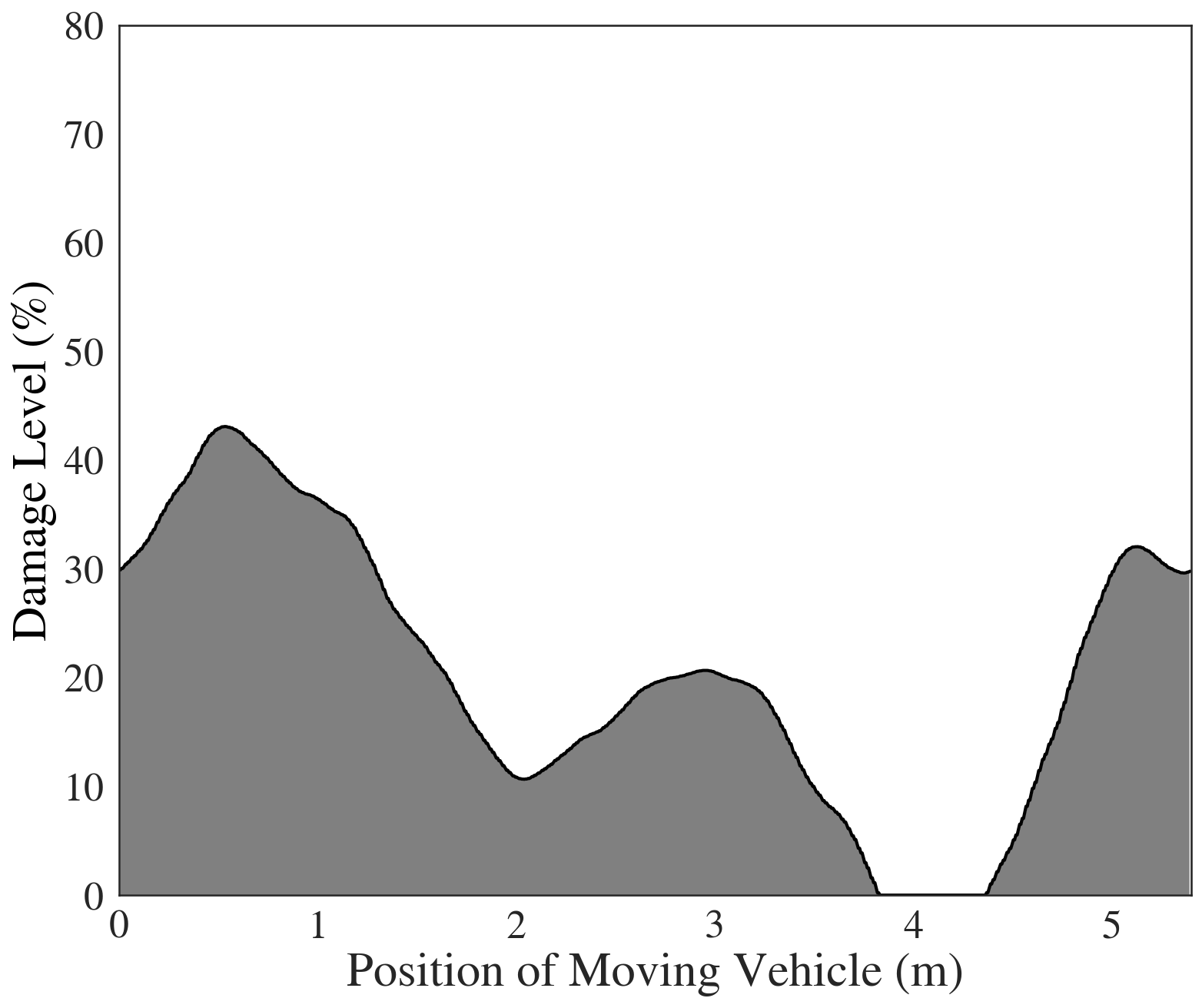} }
  \hspace{24pt}
  \subfloat[Displacement responses from forward VINO and errors\label{fig:numfwd-b} ]{\includegraphics[width=0.4\linewidth]{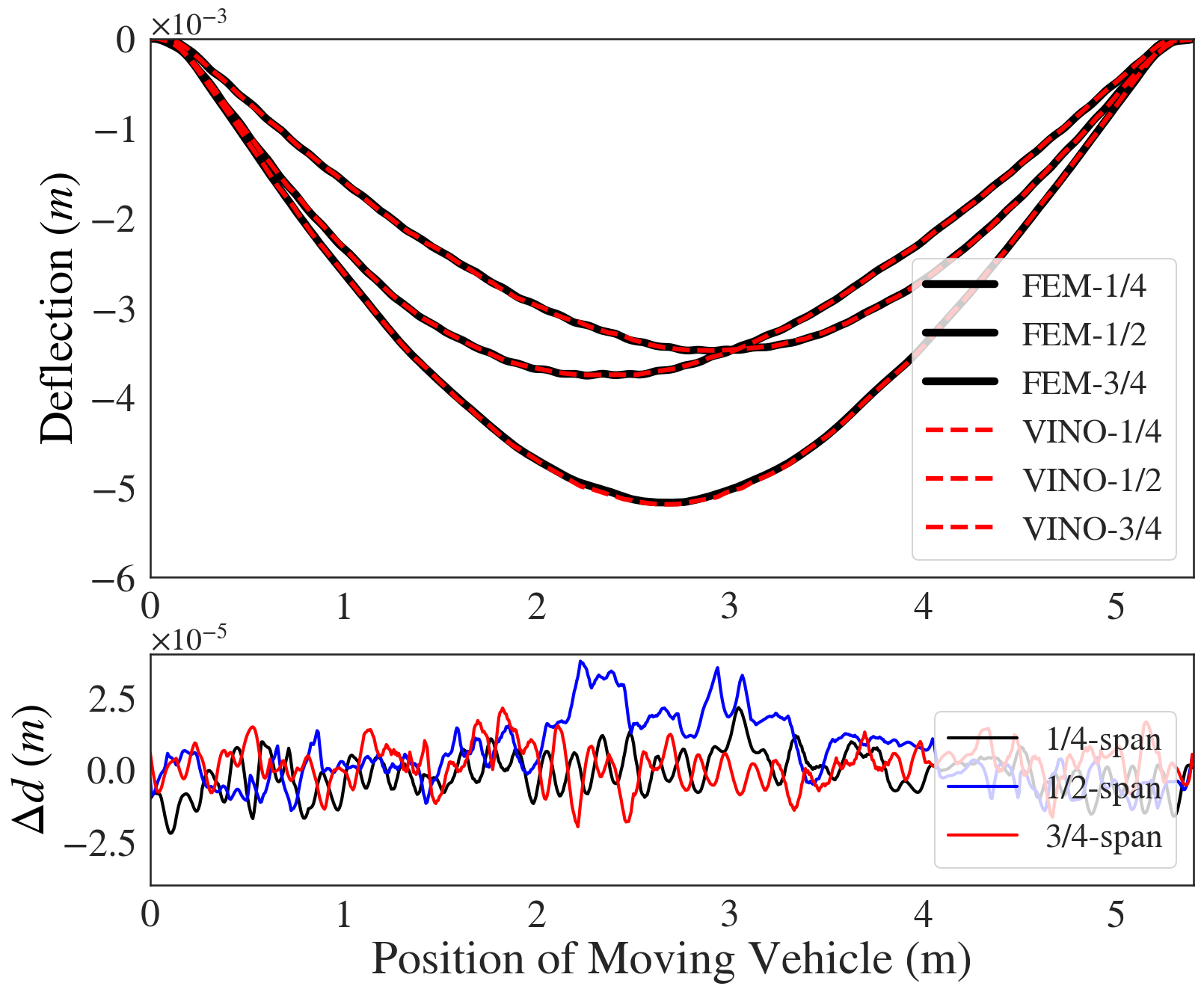}}
  \qquad
  \subfloat[Rotational angle responses from forward VINO and errors\label{fig:numfwd-c}]{\includegraphics[width=0.4\linewidth]{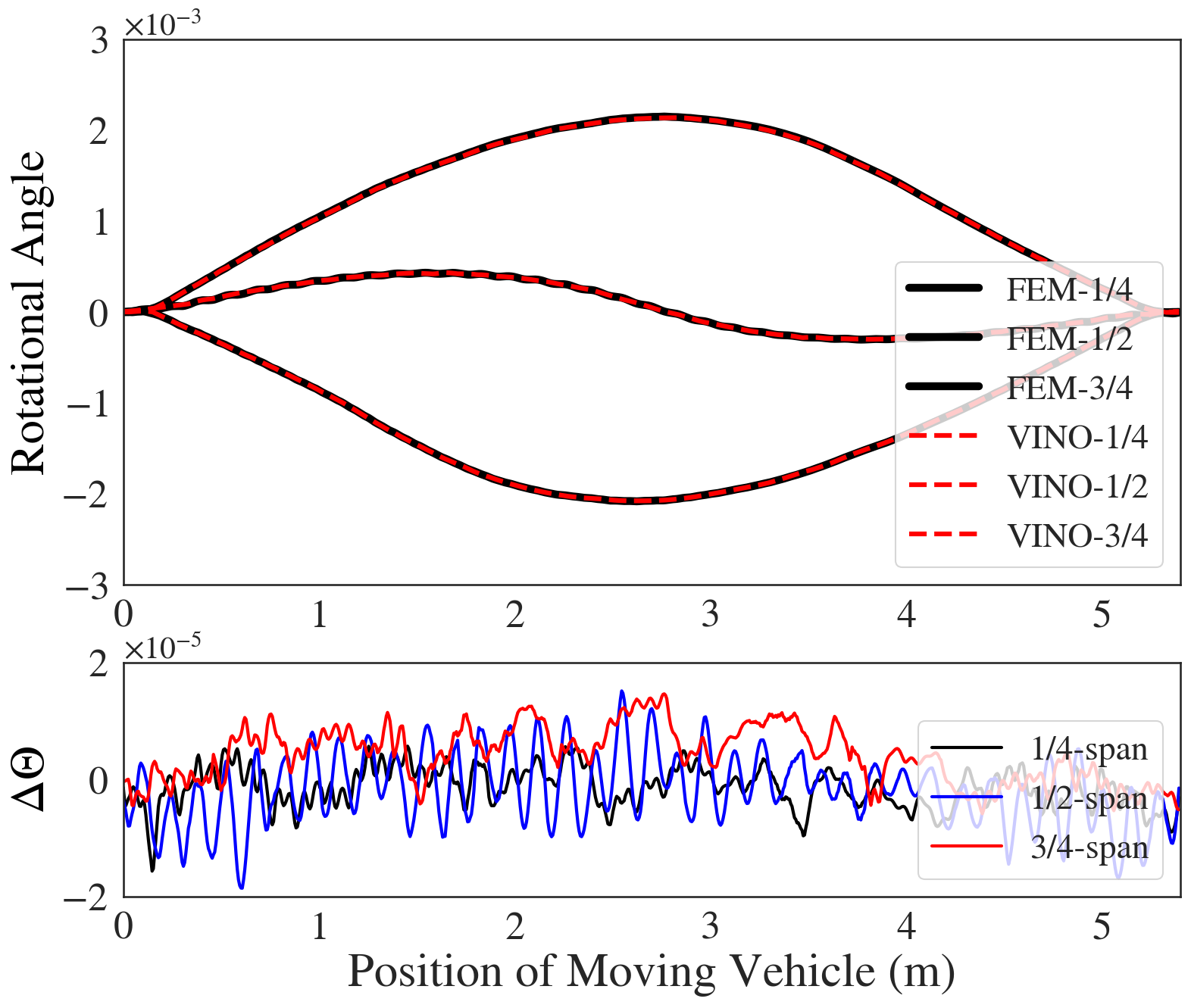} }
  \hspace{24pt}
  \subfloat[Acceleration response from forward VINO and errors\label{fig:numfwd-d}]{\includegraphics[width=0.4\linewidth]{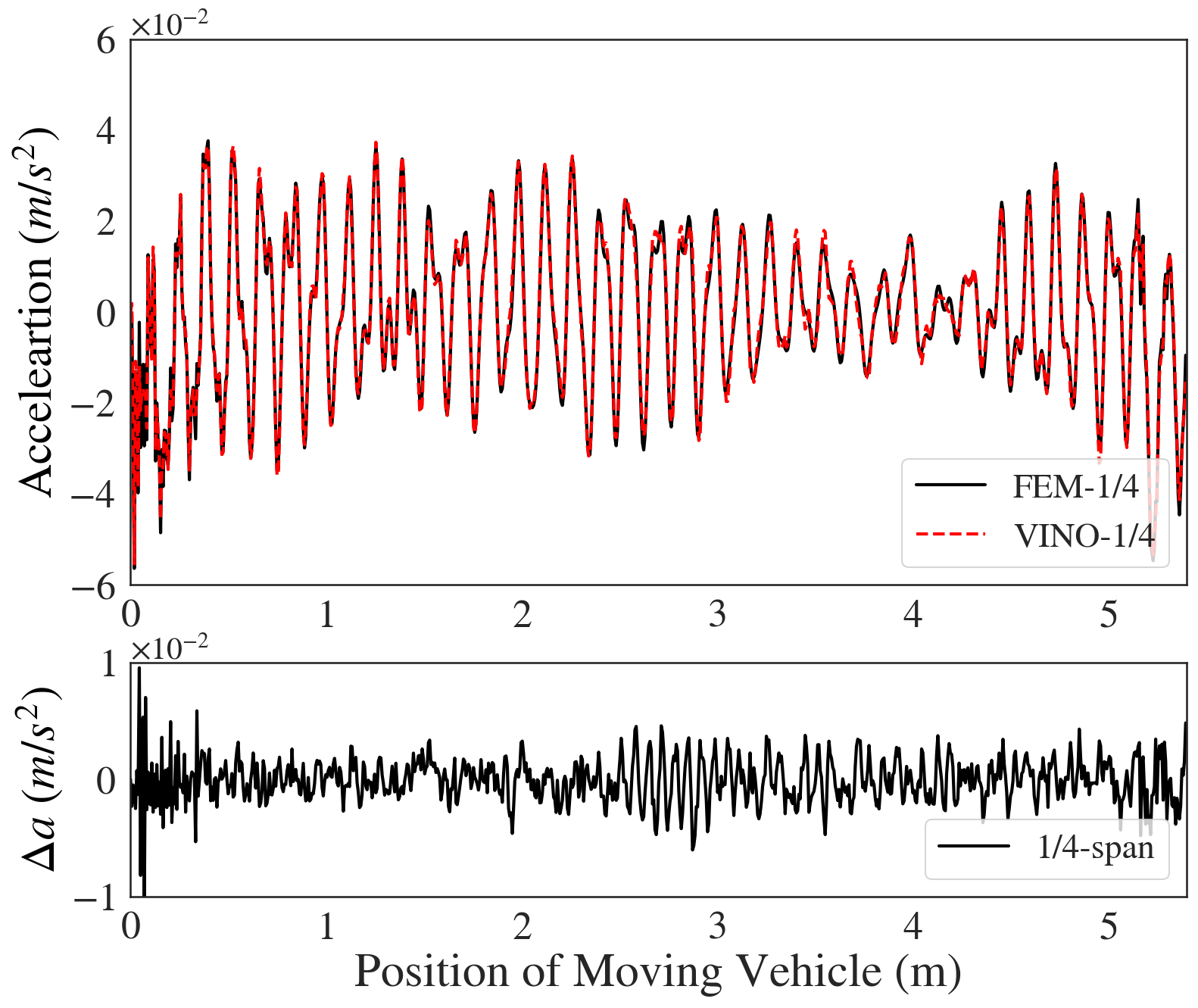}}
  \caption{Numerical Results of VINO on forward problem (a) Damage field input (b) Deflection responses and their differences between two models (c) Rotational angle responses and their differences between two model (d) Acceleration responses and their differences between two model}
  \label{fig:numerical-forward} 
\end{figure*}

VINO and the FE model should be compared to assess which is more effective, taking into account errors and computational time. Therefore, this verification aims to discuss a more efficient model for further forward problem purposes. This paper provided numerical verification for VINO models trained to generate displacement, rotational angle, and acceleration from a random damage field in Figure \ref{fig:numerical-forward}. 

To obtain displacement response, VINO was trained to predict a single channel of displacement measurement data. Therefore, three VINO models were trained to capture the response from three locations (1/4 span, mid-span, and 3/4 span). After being trained on the 1,000 training data in the VBI-FE dataset, VINO models are capable of generating the same output as the FE simulation results. Figure \ref{fig:numfwd-a} shows the damage field input to both models, while responses in Figures \ref{fig:numfwd-b} are the output of the models. As observed in Figure \ref{fig:numfwd-b}, each displacement response was obtained from each VINO with an error below 40 $\mu m$. In addition, the FE simulation time is 66 seconds to obtain response at 1028 nodes on average based on a single core of Intel(R) Core(TM) i9-10850K CPU @3.6GHz with NVIDIA GeForce RTX 3090. In comparison, after the training process is complete, the inference time of VINO is only 34 $ms$ to obtain one response based on a single Nvidia V100 GPU. This implies that for the full simulations, Fourier Neural Operator will be 19 times faster than the FE model once it is trained. For a specific task that requires only selected nodes, VINO will achieve notably higher computational efficiency in structural response prediction because of the efficient utilization of parallel computing resources of GPU accelerators and there is no need to form a stiffness matrix or matrix solver in VINO. Therefore, the result can be concluded for the excellent performance of the VINO in learning to map the damage field to the displacement at an arbitrary node, which is at least as accurate as the FE model.

Figure \ref{fig:numfwd-c} shows the performance of VINO in fitting the rotation angle of the bridge. Similar to the displacement responses, the rotational angles of bridge responses from VINO were considerably close to those obtained from the FE model. The maximum error of rotation angle is less than $10^{-5}$ degrees observed at the bottom of Figure \ref{fig:numfwd-c}. The inference time of VINO is similar to that of the displacement prediction model and is notably faster than the FE model. This suggests that VINO could reproduce FE simulation results of rotational angle. 

The acceleration response data vary and fluctuate over time domain and are more sensitive to FE modeling parameters such as damping ratio. In Figure \ref{fig:numfwd-d}, the acceleration responses of the quarter-span generated from VINO and FE simulation were shown. Although higher errors were observed in acceleration prediction, the general trend of acceleration obtained from FE simulation is replicated by the VINO model. The error between VINO and FE simulation was lower than 1 $mm\cdot s^{-2}$. After the training process is complete, like displacement and rotational angle, the acceleration response at each node was inferred within 35 $ms$, suggesting a higher efficiency than the FE model. The Fourier mode in the Fourier layers of VINO was 16 as applied to approximate the one-dimensional field in \citet{zongyi}, which was discussed to be sufficient in the approximation. Although the current progress of this study does not aim to investigate the most efficient Fourier mode, it should be noted that an analysis of the Fourier mode may be needed to optimize the computational time and accuracy.

As a result, all three numerical verifications concluded that VINO has the competence to simulate the deflection, rotational angle, and acceleration responses as the FE model created in the forward problem at the quarter, mid, and three-quarter span locations. This further suggests the ability to generate other structural responses of the other nodes on the bridge in the dataset. From the results, an apparent advantage of the VINO over the FE simulation is the inference time, which is almost 2000\% faster than the FE model for a full simulation. Another suggestion for the full operation is the two-dimensional VINO to predict the displacement field as a function of coordinate ($x$) and time ($t$), which should be investigated for applications in forward VBI problems. It is believed that if the two-dimensional model provides high accuracy, it can be a great benefit to future VBI research.

\subsection{Inverse VINO for Structural Health Monitoring}

\begin{figure*}[ht!] 
\centering
  \subfloat[Displacement responses as inputs to inverse VINO\label{fig:numinv-a}]{\includegraphics[width=0.4\linewidth]{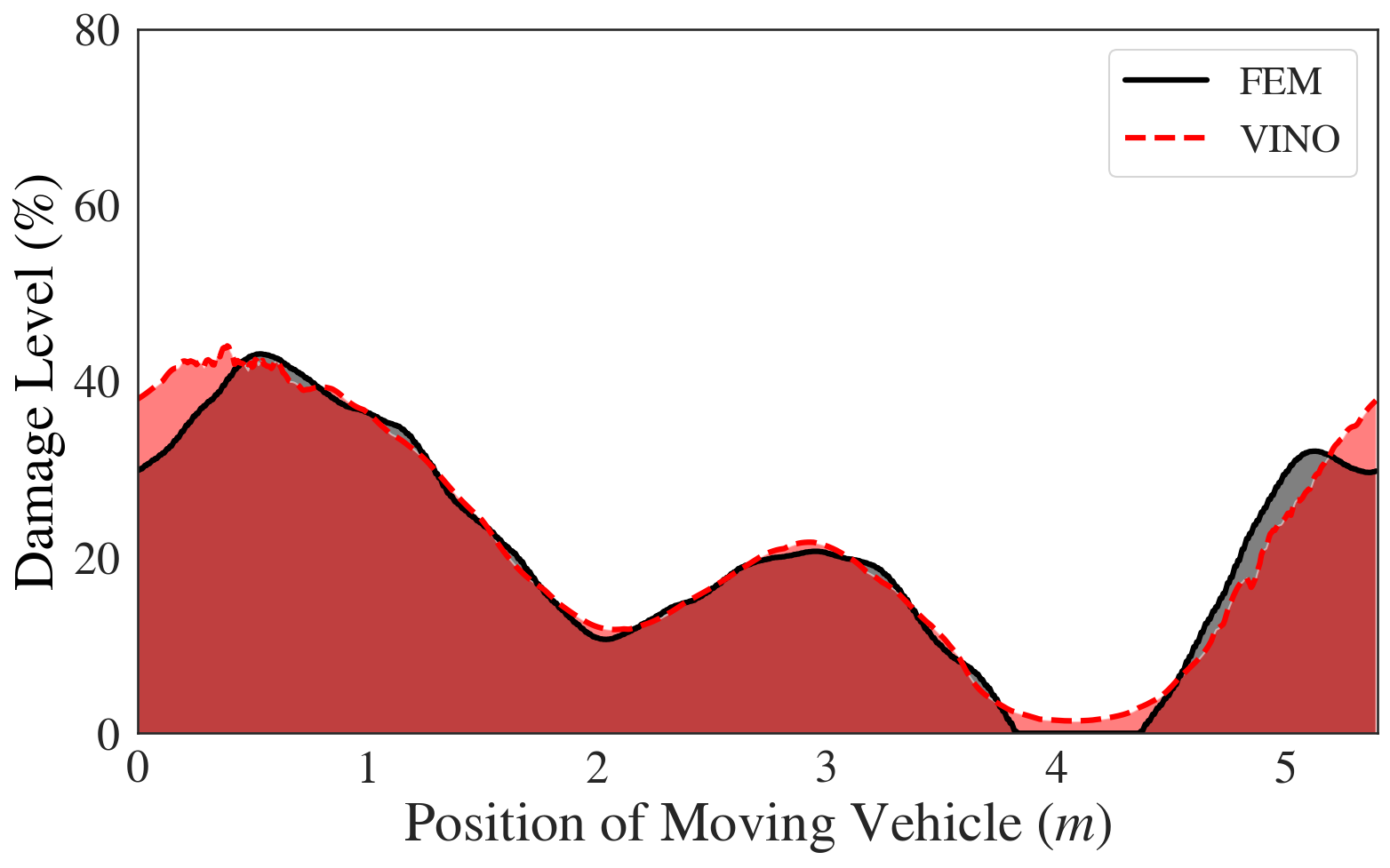}}
  \hspace{24pt}
  \subfloat[The damage field outputted from inverse VINO\label{fig:numinv-b}]{\includegraphics[width=0.4\linewidth]{figures/numerical-inverse/invdef2.png}}
  \qquad
  \subfloat[Acceleration responses as inputs to inverse VINO\label{fig:numinv-c}]{\includegraphics[width=0.4\linewidth]{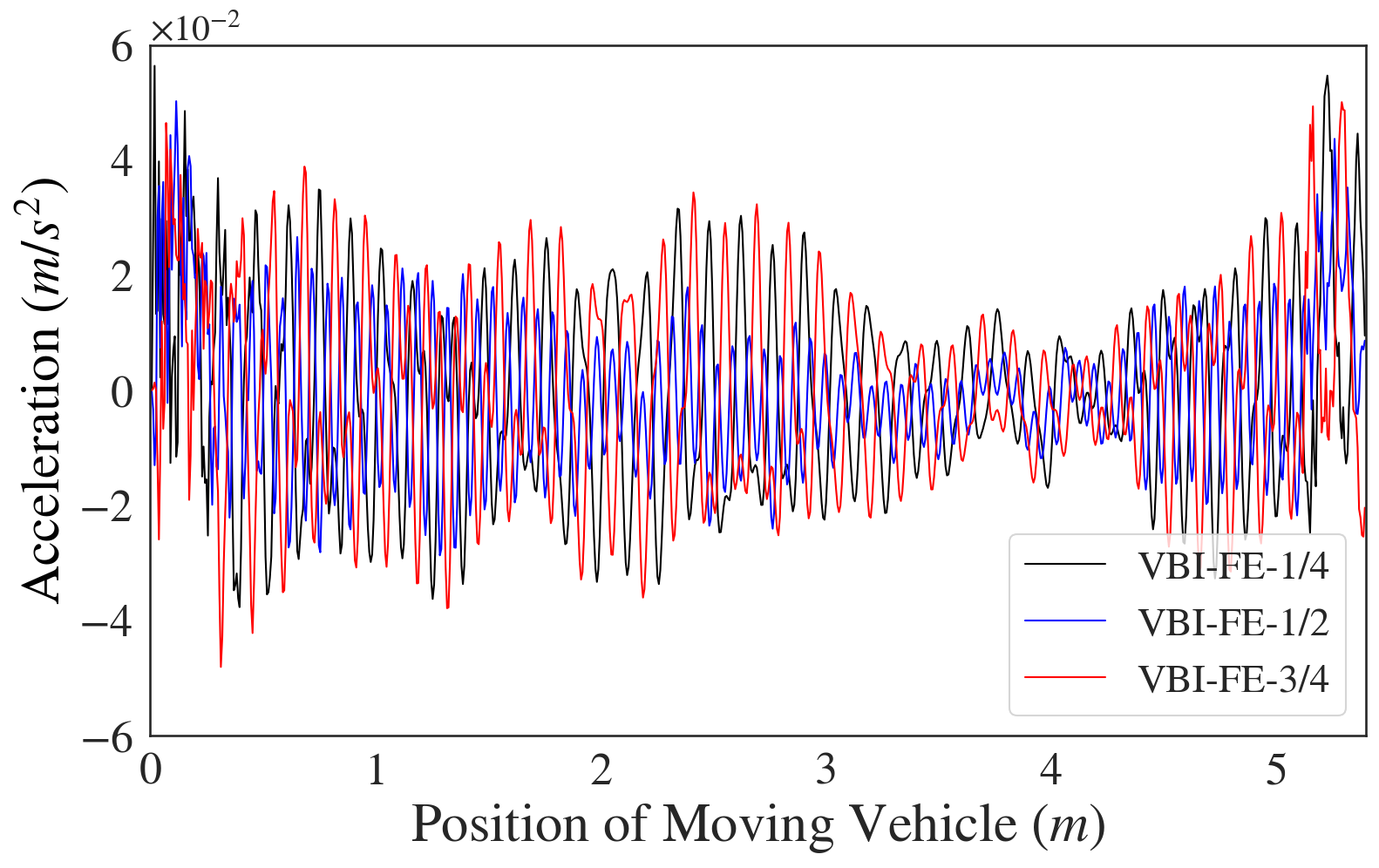} }
  \hspace{24pt}
  \subfloat[The damage field outputted from inverse VINO\label{fig:numinv-d}]{\includegraphics[width=0.4\linewidth]{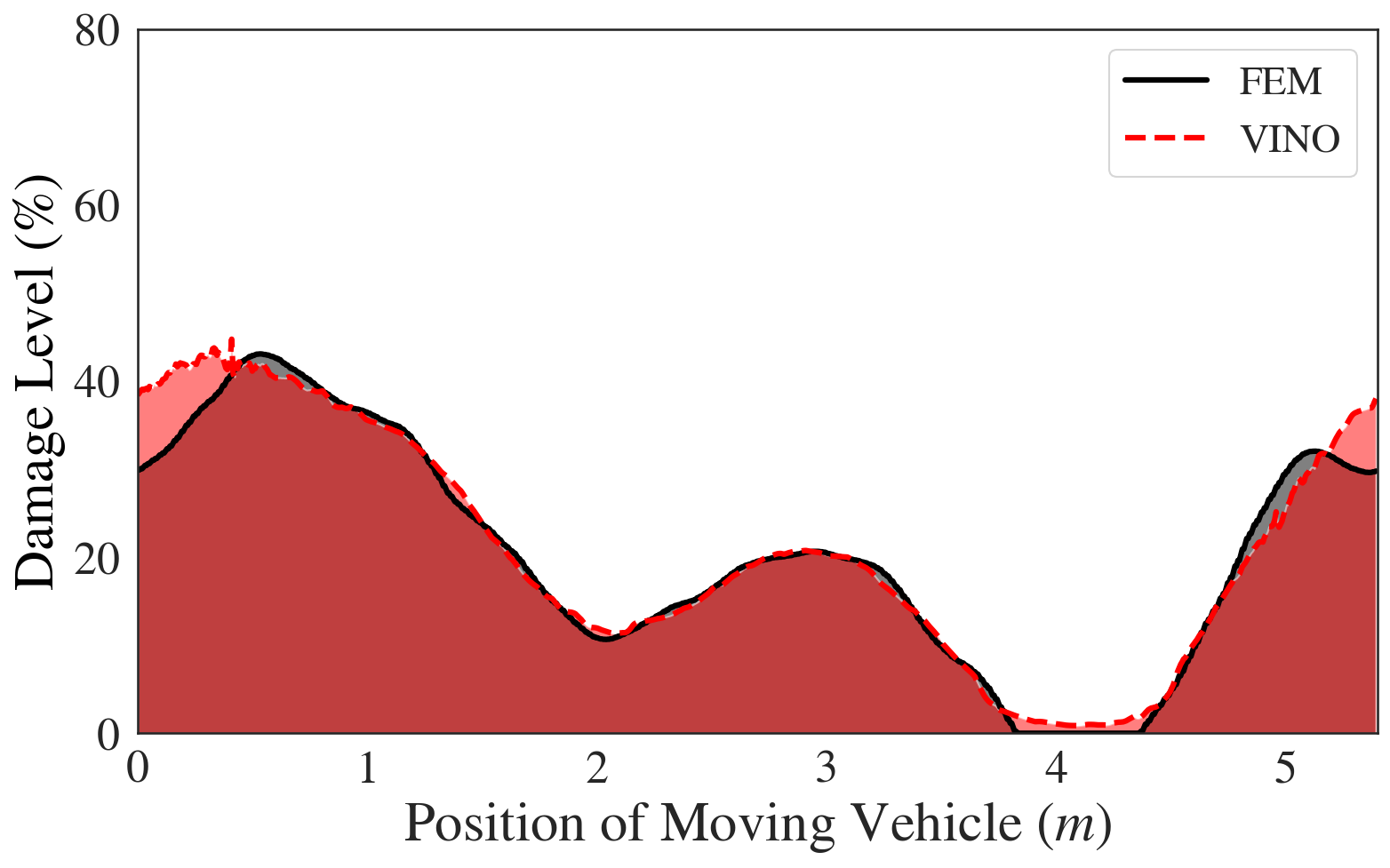}}
  \caption{Numerical Results of VINO on inverse problem (a) Deflection responses at the quarter, half, three-quarter span (b) Theoretical and Predicted damage fields from by VINO (c) Acceleration responses at the quarter, half, three-quarter span (d) Theoretical and Predicted damage fields from by VINO}
  \label{fig:numerical-inverse} 
\end{figure*}

As mentioned in section \ref{sec:introduction}, damage detection in BHM is considered an inverse problem because the approach aims to find the causal parameter (damage distribution) from the detected vibration responses. In this case, the damage on the bridge is the parameter that causes the difference in the measurement of structural responses. This section presents the results of the VINO that maps from the structural responses field to the damage field. The inputs of VINO are structural responses at the quarter, mid, and three-quarter spans on a numerical bridge, and the output is the damage field. In this paper, only two responses, as shown in Figure \ref{fig:numerical-inverse} — displacement and accelerations were investigated because these can be easily measured with displacement sensors and accelerometers.

To predict damage fields from displacement responses, the VINO was trained by displacements (Figure \ref{fig:numinv-a}) at three nodes, including quarter, half, and three-quarter spans, and the VINO output was the damage field. In Figure \ref{fig:numinv-b}, the black line depicts the theoretical damage field, which is an input to the FE model to generate the structural responses, and the red line represents the damage field predicted by VINO. Therefore, numerical results from Figure \ref{fig:numinv-b} showed the considerably accurate prediction of VINO in mapping between displacement and damage fields, suggesting the capability to predict damage along the bridge span. The inaccurate prediction can be seen at the bridge's start and end because the structural response is not sensitive to the possible damage at the beam end.

Although acceleration responses have more high-frequency components compared to the deflection, as shown in Figure \ref{fig:numinv-c}, the damage fields in Figure \ref{fig:numinv-d} predicted by VINO trained by the acceleration inputs showed results similar to the damage fields predicted from the displacement inputs. Three acceleration responses at the quarter, mid, and three-quarter nodes were the inputs of the VINO model, which was able to predict an accurate damage field in the numerical verification. This suggested investigations of VINO for the experimental data with only sensors at the three locations. 

Observations from the simulation data-based investigation demonstrated that it is sufficient to draw a conclusion that VINO is an excellent candidate in damage detection for determination, localization, and quantification. Because the encoder-based architecture maps between two function spaces, the machine learning model can predict the damage field from displacement or acceleration responses discussed above in Figure \ref{fig:numerical-inverse}. Similar to several other research, these numerical verifications, in fact, do not provide sufficient justification for the practicality of the model. To overcome the challenges of data-driven SHM mentioned in review articles \citep{azimi, toh, avci, malekloo, meisam}, the numerically-verified Fourier Neural Operator will be validated with the experimental data in the next section \ref{sec:ExpVal} as mentioned in the framework Figure \ref{fig:framework}. In addition, it is believed that the one-dimensional FNO suits better with this inverse problem in comparison to the two-dimensional FNO because the inputs can be detected responses from as many sensors as necessary and the output is a damage field, which is strictly one dimension. However, if the problem goes up to the three-dimensional structure, it is suggested to further develop the higher-dimensional VINO models. 

\section{Performance of Fine-tuned VINO on the VBI-EXP dataset}
\label{sec:ExpVal}

In this section, the experimental validation for VINO was reported for both forward and inverse problems. In the forward problem of structural simulation, this validation aims to provide a justification that after transfer learning, VINO can generate more reliable structural responses than the conventional FE model. For the inverse problem, validation is provided to demonstrate data-driven BHM. To approach those goals as a practical approach in the forward and inverse problems, the model must be validated by the data of the actual bridge at the damage state (not the numerical data) in which the model is making a prediction. This means the model will be pre-trained on the VBI-FE dataset and fine-tuned on the VBI-EXP dataset or the real experimental data from the bridge only at the healthy state by transfer learning. Subsequently, VINO can predict the damage fields on the bridge when the damages appear.

\subsection{Fine-tuned Forward VINO for Structural Simulation}

In VBI problems, it is still a challenging problem to generate the exact responses as the experiments, even though the numerical and experimental setups are identical due to the fact that the numerical models (FE method) have some assumptions (elastic material, Rayleigh damping model, etc). In machine learning-based approaches, transfer learning approaches are applied to a model to utilize an understanding of the pattern or function of a specific task.

\begin{figure*}[ht!] 
\centering
  \subfloat[Displacement responses of INT scenario\label{fig:expfwd-a}]{\includegraphics[width=0.4\linewidth]{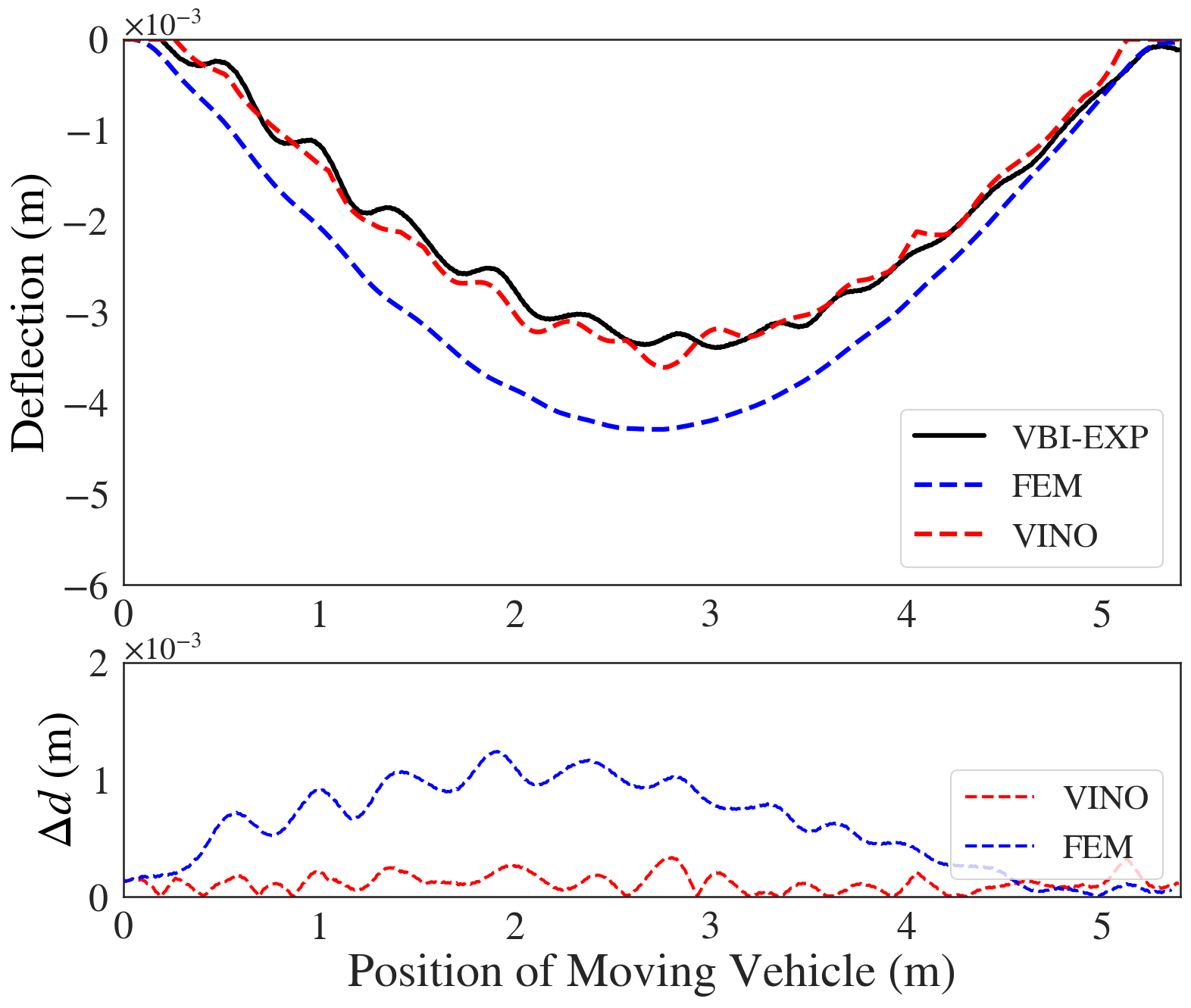}}
  \hspace{24pt}
  \subfloat[Displacement responses of DMG1 scenario\label{fig:expfwd-b}]{\includegraphics[width=0.4\linewidth]{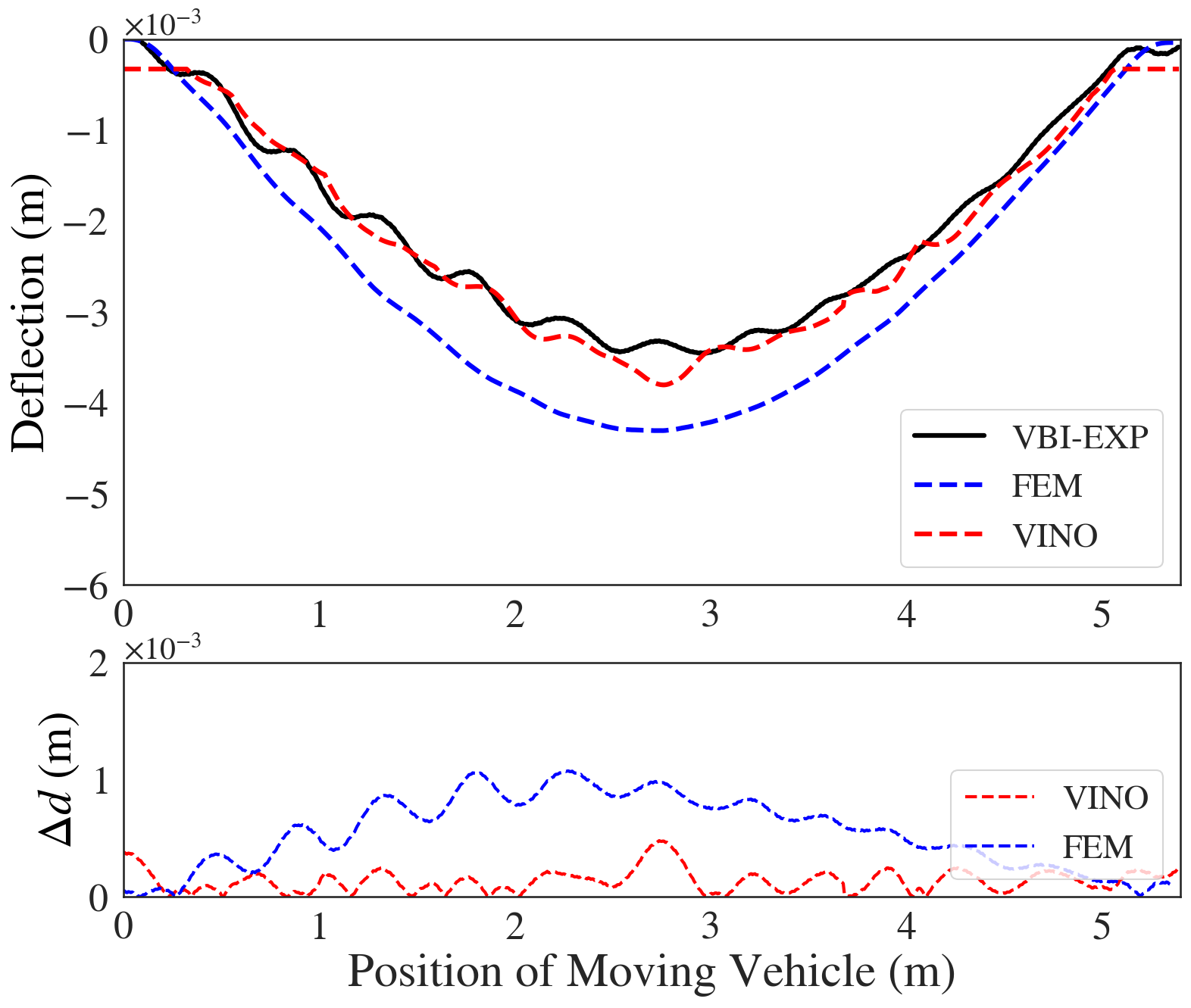}}
  \qquad
  \subfloat[Displacement responses of DMG2 scenario\label{fig:expfwd-c}]{\includegraphics[width=0.4\linewidth]{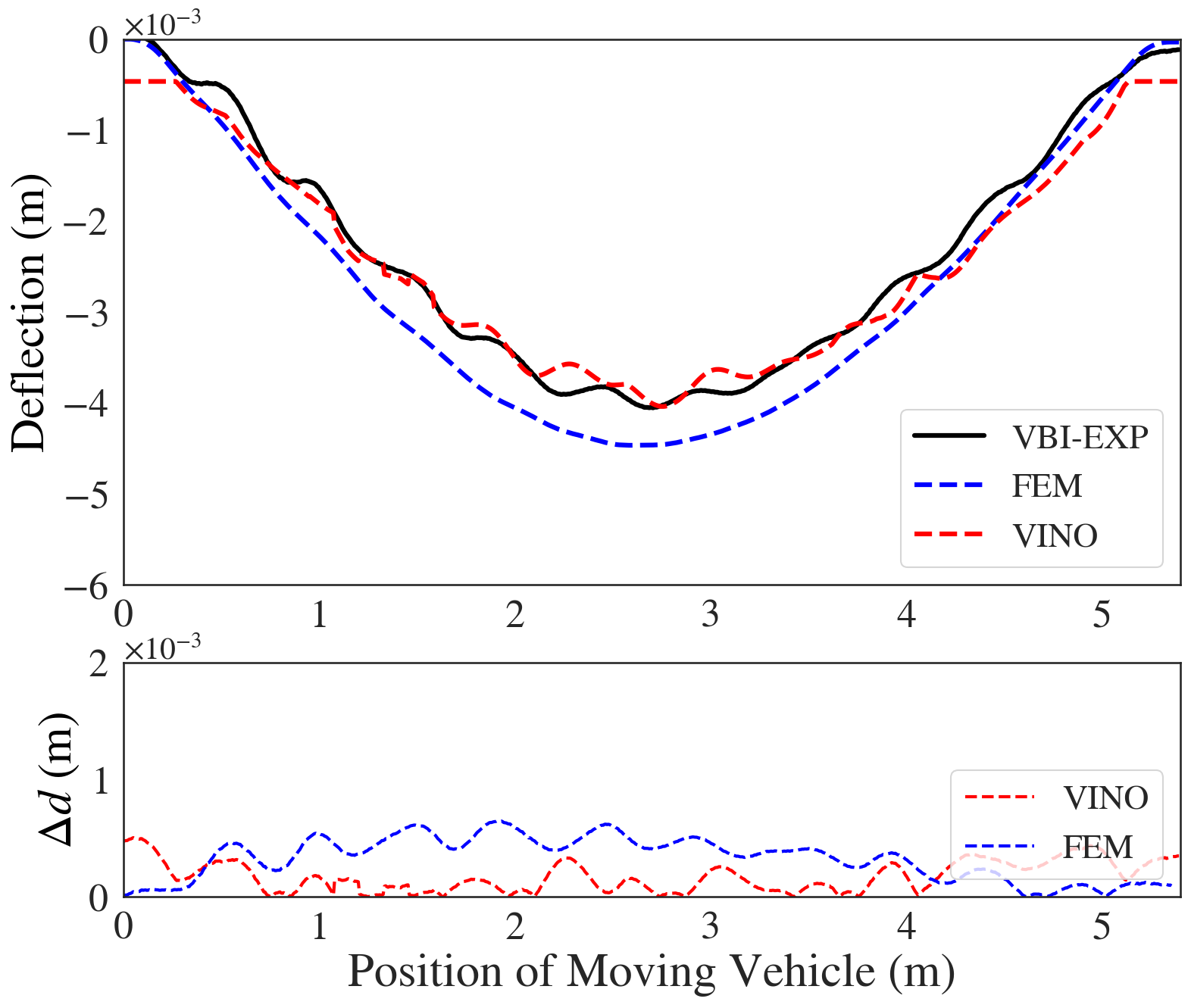}}
  \hspace{24pt}
  \subfloat[Displacement responses of DMG3 scenario\label{fig:expfwd-d}]{\includegraphics[width=0.4\linewidth]{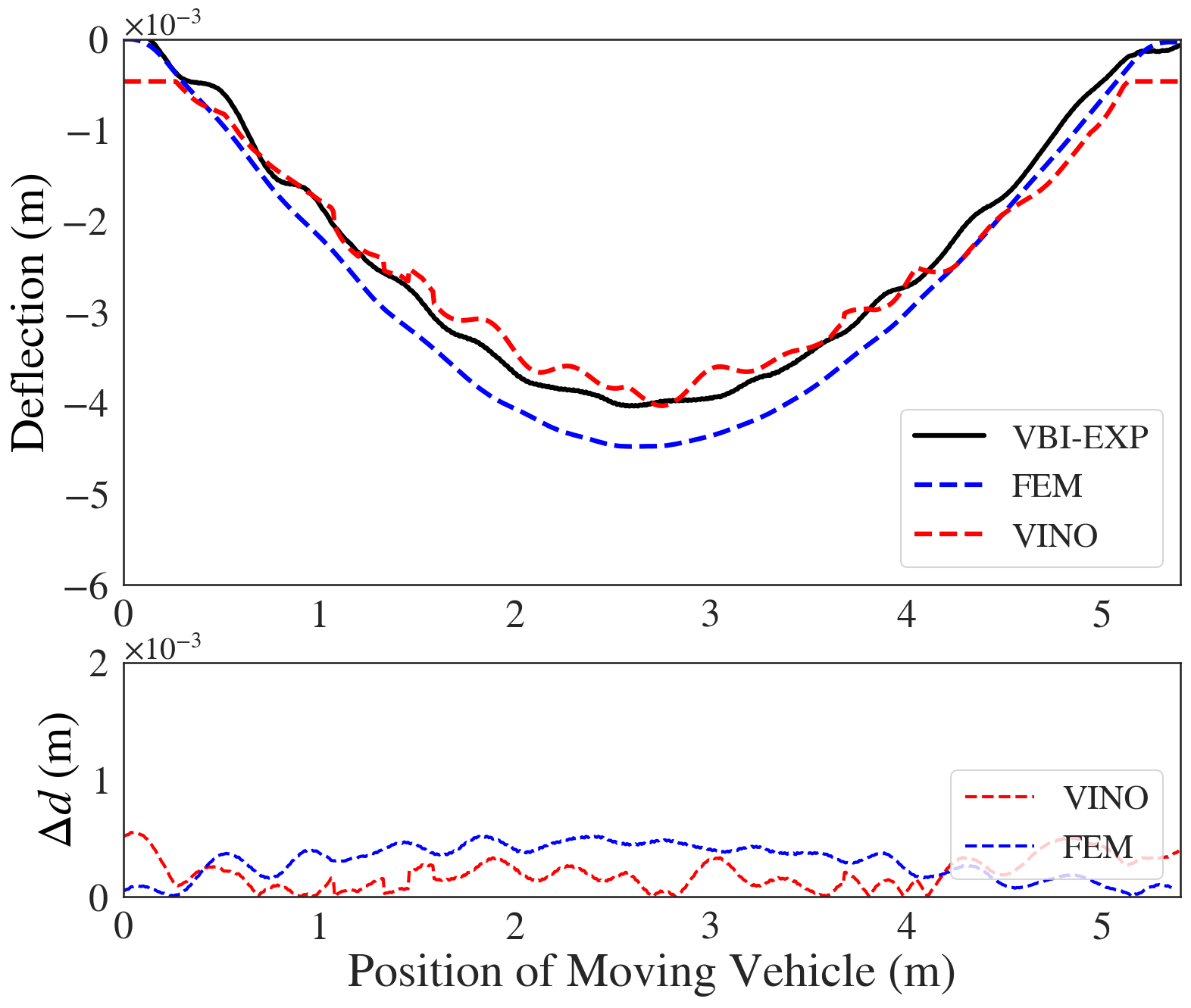}}
  \caption{Experimental Validation of Fourier Neural Operator on Forward Problem. These are deflection responses comparison of (a) INT (b) DMG1 (c) DMG2 (d) DMG3 scenarios}
  \label{fig:experimental-forward}  
\end{figure*}

In the structural simulation problems, the forward VINO was fine-tuned to simulate displacement responses only. After the VINO model was pre-trained on the VBI-FE dataset of 1,000 simulations and validated in section \ref{sec:NumVal}, it was then fine-tuned on the VBI-EXP dataset obtained from experiments on the laboratory bridge. As mentioned in previous sections, only the responses from the bridge at the healthy state (INT scenario) were adopted to train the last two layers of the VINO model, while other layers in the model were frozen. After that, the fine-tuned VINO model was further validated by three damage scenarios as prepared in the VBI-EXP dataset: DMG1, DMG2, and DMG3 as reported in the experimental setup in the section \ref{sec:NumSetup}.

Figure \ref{fig:experimental-forward} showed the comparison between the collected experimental data in the VBI-EXP dataset, FE simulation, and the VINO model trained on the VBI-FE dataset and fine-tuned on the VBI-EXP dataset in simulating the displacement responses of the bridge. Figure \ref{fig:expfwd-a} is the results on the healthy bridge (i.e. training set in the VBI-EXP dataset). The difference between the actual experimental responses and the FE model of the healthy bridge is around 1.25 $mm$ at the maximum experimental displacement (3.4 $mm$), equivalent to a relative error of 37 percent. However, the error between the VINO and experimental results is approximately 0.3 $mm$ along the response field. As the pre-trained VINO was fine-tuned with the INT scenario, a small error was expected. A similar trend was shown in Figure \ref{fig:expfwd-b} for the DMG1 scenario. Compared to experimental data, the FE model and VINO showed maximum errors of nearly 1.1 $mm$ and 0.5 $mm$, respectively. The performance of the FE model for simulating displacement responses under DMG2 and DMG3 in Figures \ref{fig:expfwd-c} and \ref{fig:expfwd-d} showed a larger error compared to VINO. The maximum error of the FE model was still above 0.65 $mm$ along the displacement responses. In comparison, the fine-tuned VINO model reduced the displacement error to less than 0.55 $mm$ along the span of the bridge. It is evident in DMG1, DMG2, and DMG3 that it was difficult for VINO to predict the response at the start and the end as the error was always one of the largest errors along the span. This error may come from the data processing during the synchronization of optical and three displacement sensors.

Therefore, it is concluded that the fine-tuned VINO showed better performance compared to the FE model in the forward problem (i.e. structural simulation problem). The maximum error between test and simulation results was reduced in VINO, and the oscillation of the actual displacement responses was also predicted by the VINO model. This may be attributed to the Rayleigh damping model adopted in FE simulation, which may dampen out high-frequency responses beyond the second frequency of the bridge structure. The fine-tuned VINO model was able to generate high-frequency component displacement responses, as shown in Figure \ref{fig:experimental-forward}.

\subsection{Fine-tuned Inverse VINO for Structural Health Monitoring}

The transfer learning of the inverse problem for BHM shows the practical application of VINO. In this section, the validation will be reported to illustrate that the proposed model VINO is able to predict DMG1, DMG2, and DMG3 after conducting the fine-tuning approach only with responses of the healthy bridge in the VBI-EXP dataset. In this paper, both displacement and acceleration responses at the 1/4-span, mid-span, and 3/4-span are used to investigate the practical application of the VINO with displacement data or acceleration data.

\begin{figure*}[ht!] 
\centering
  \subfloat[Displacement responses of INT scenario\label{fig:expinvdef-a}]{\includegraphics[width=0.35\linewidth]{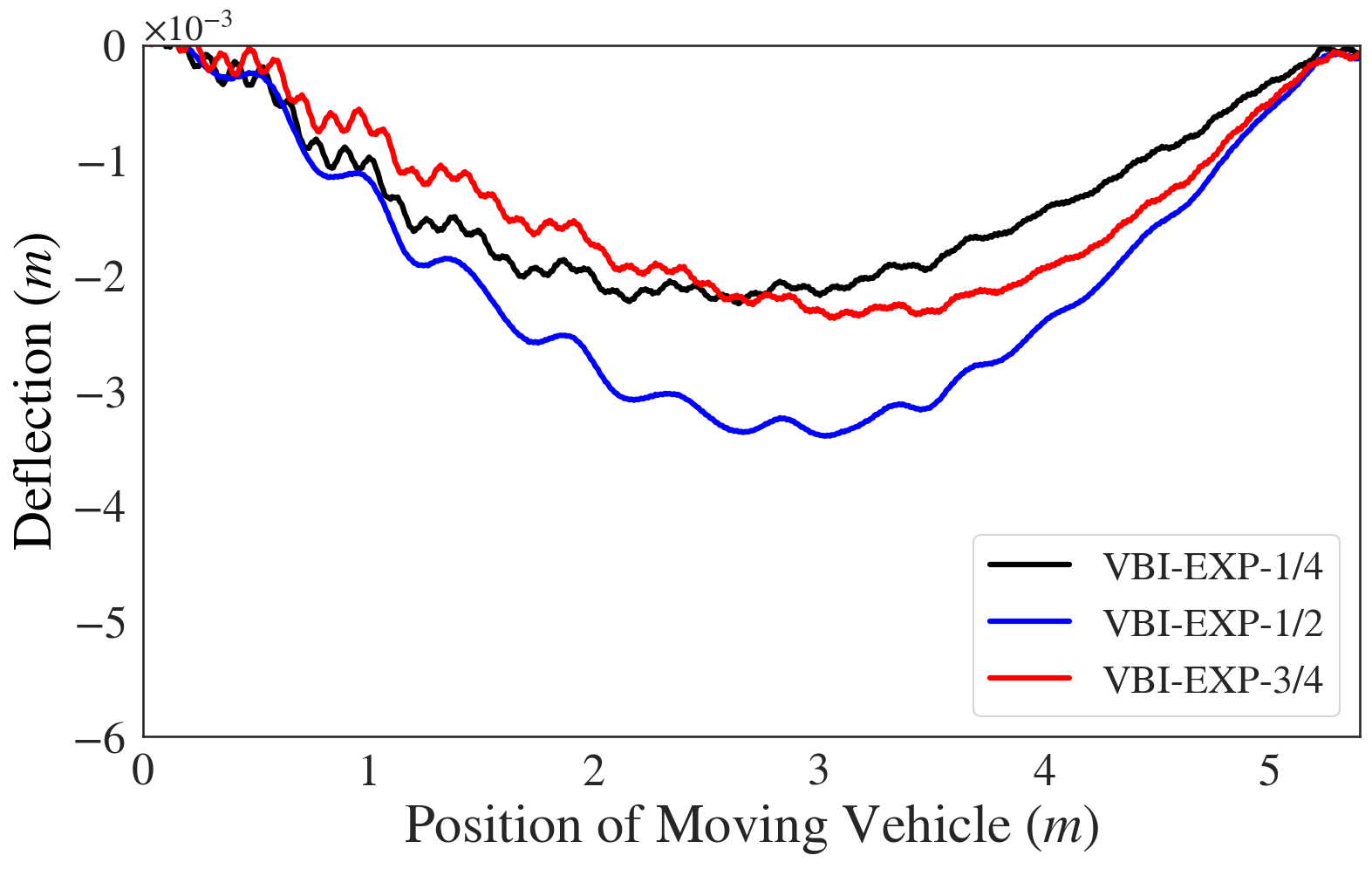}}
  \hspace{24pt}
  \subfloat[Predicted damage field of INT scenario\label{fig:expinvdef-b}]{\includegraphics[width=0.35\linewidth]{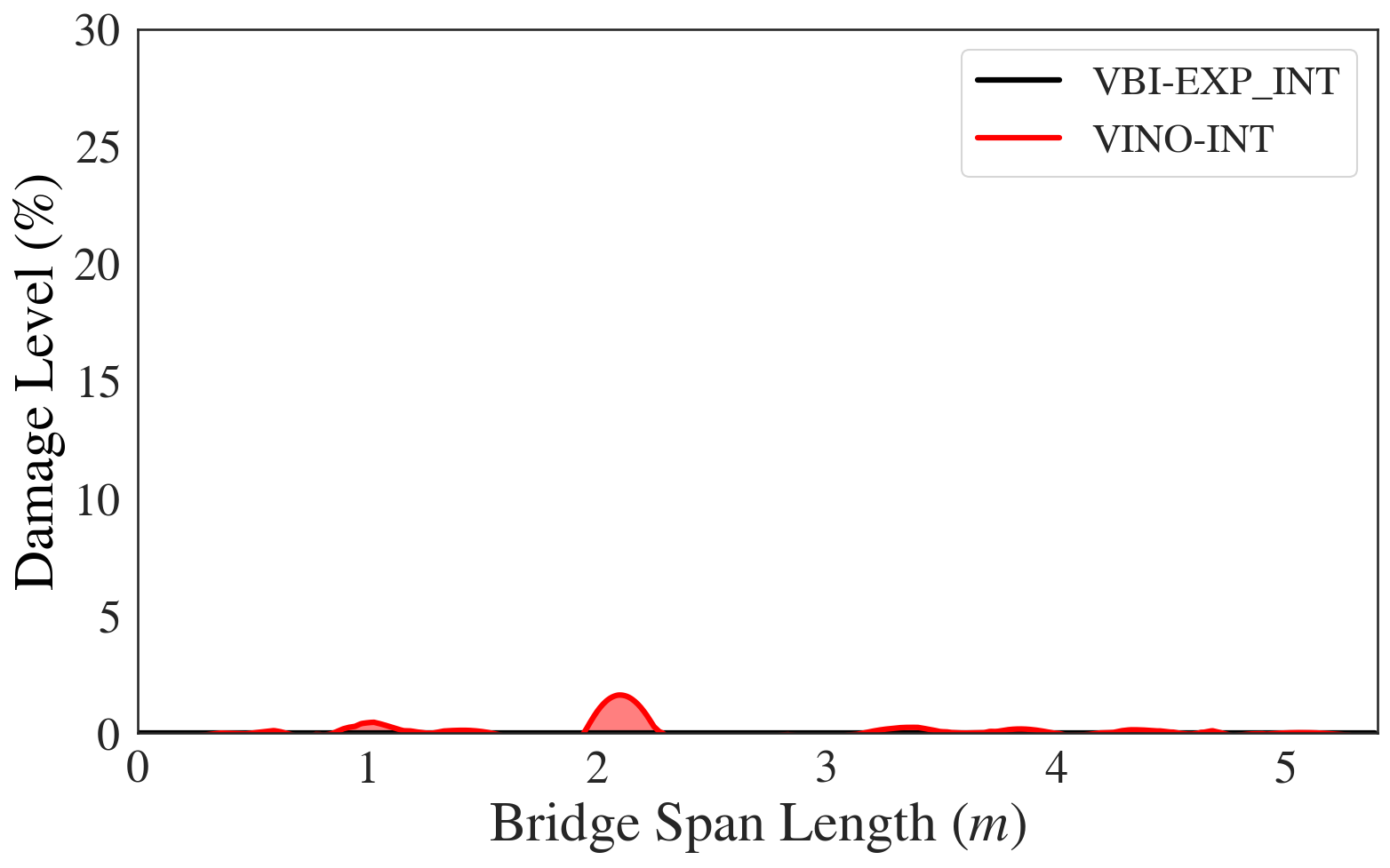}}
  \qquad
  \subfloat[Displacement responses of DMG1 scenario\label{fig:expinvdef-c}]{\includegraphics[width=0.35\linewidth]{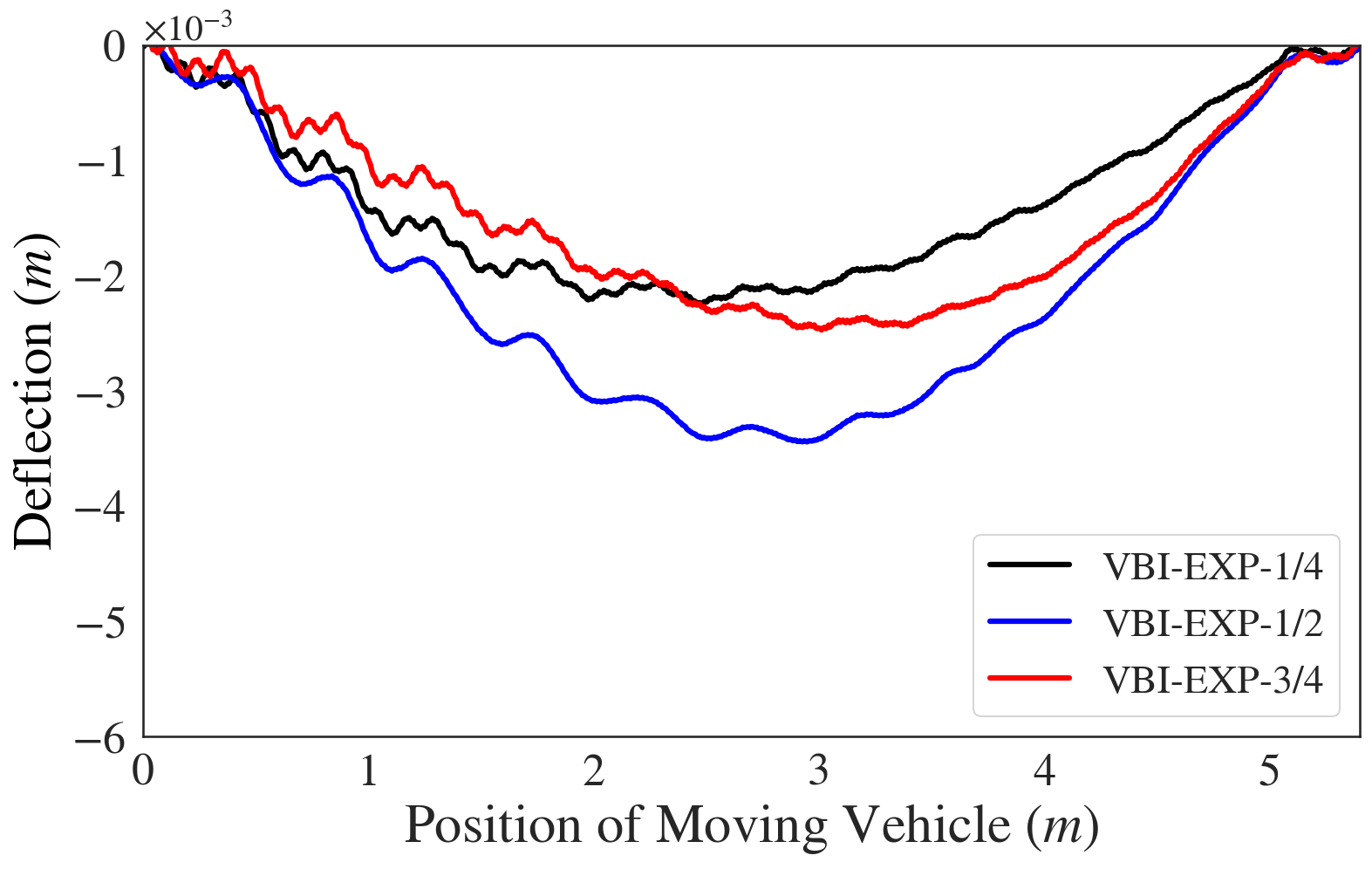}}
  \hspace{24pt}
  \subfloat[Predicted damage field of DMG1 scenario\label{fig:expinvdef-d}]{\includegraphics[width=0.35\linewidth]{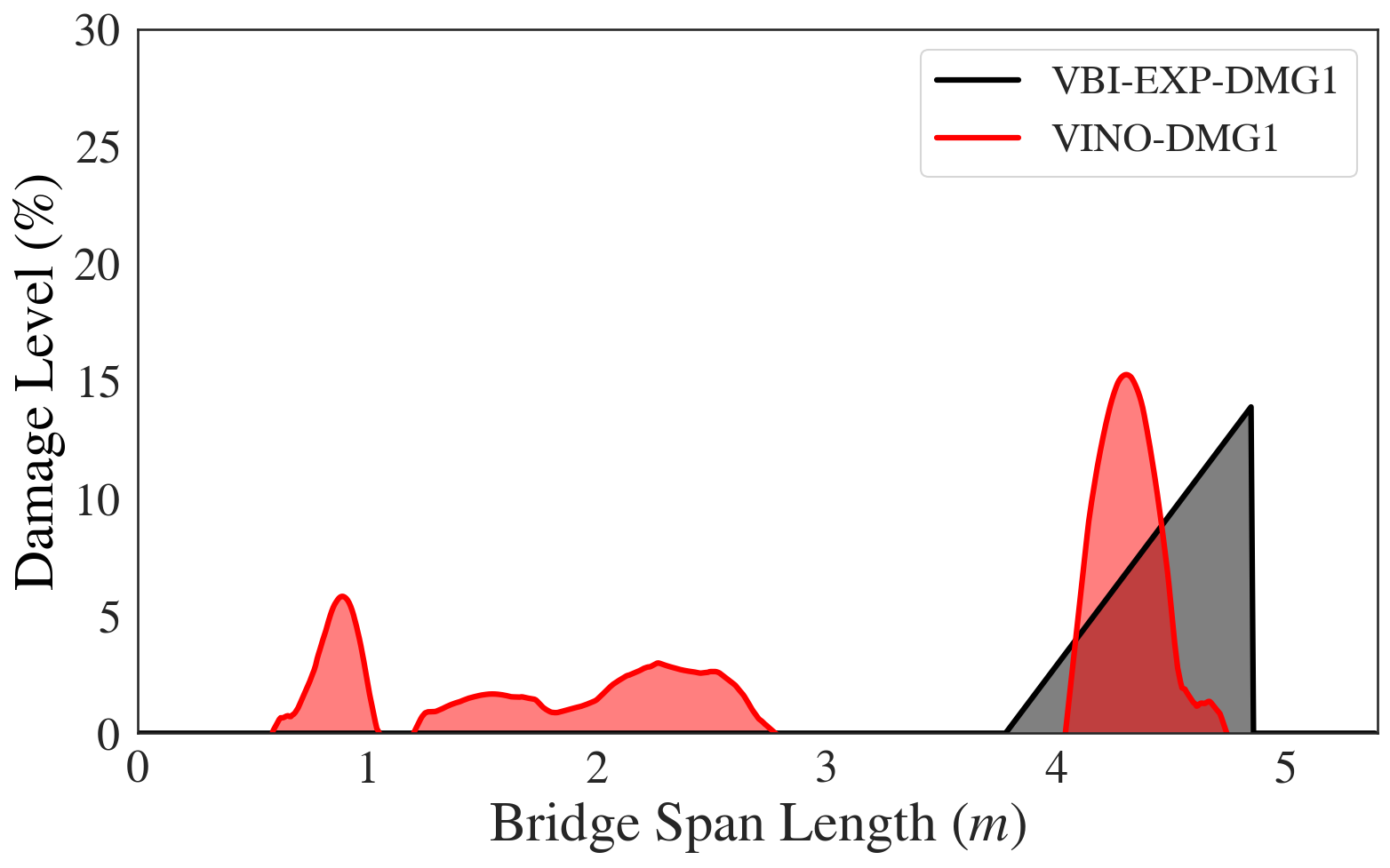}}
  \qquad
  \subfloat[Displacement responses of DMG2 scenario\label{fig:expinvdef-e}]{\includegraphics[width=0.35\linewidth]{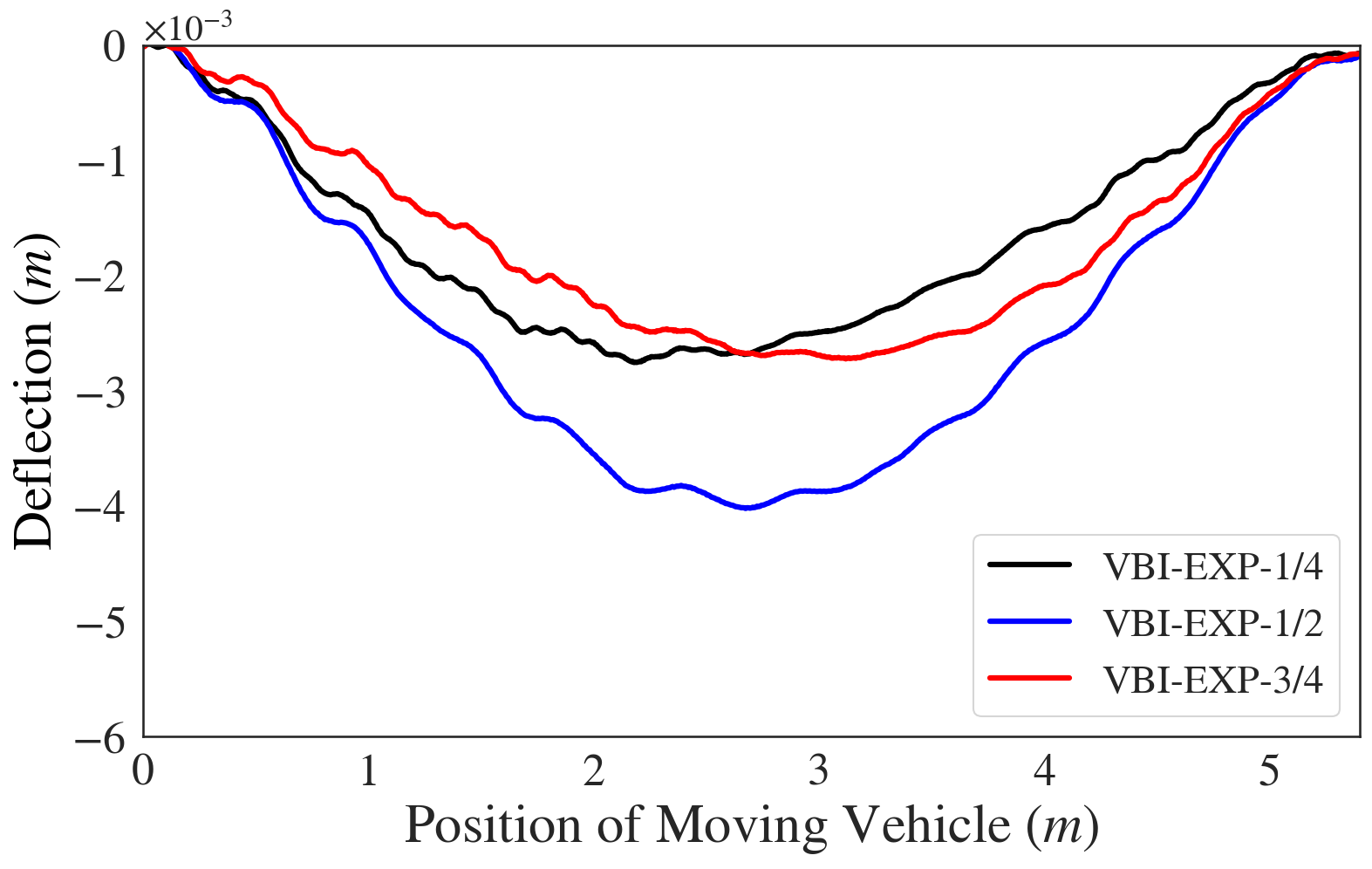}}
  \hspace{24pt}
  \subfloat[Predicted damage field of DMG2 scenario\label{fig:expinvdef-f}]{\includegraphics[width=0.35\linewidth]{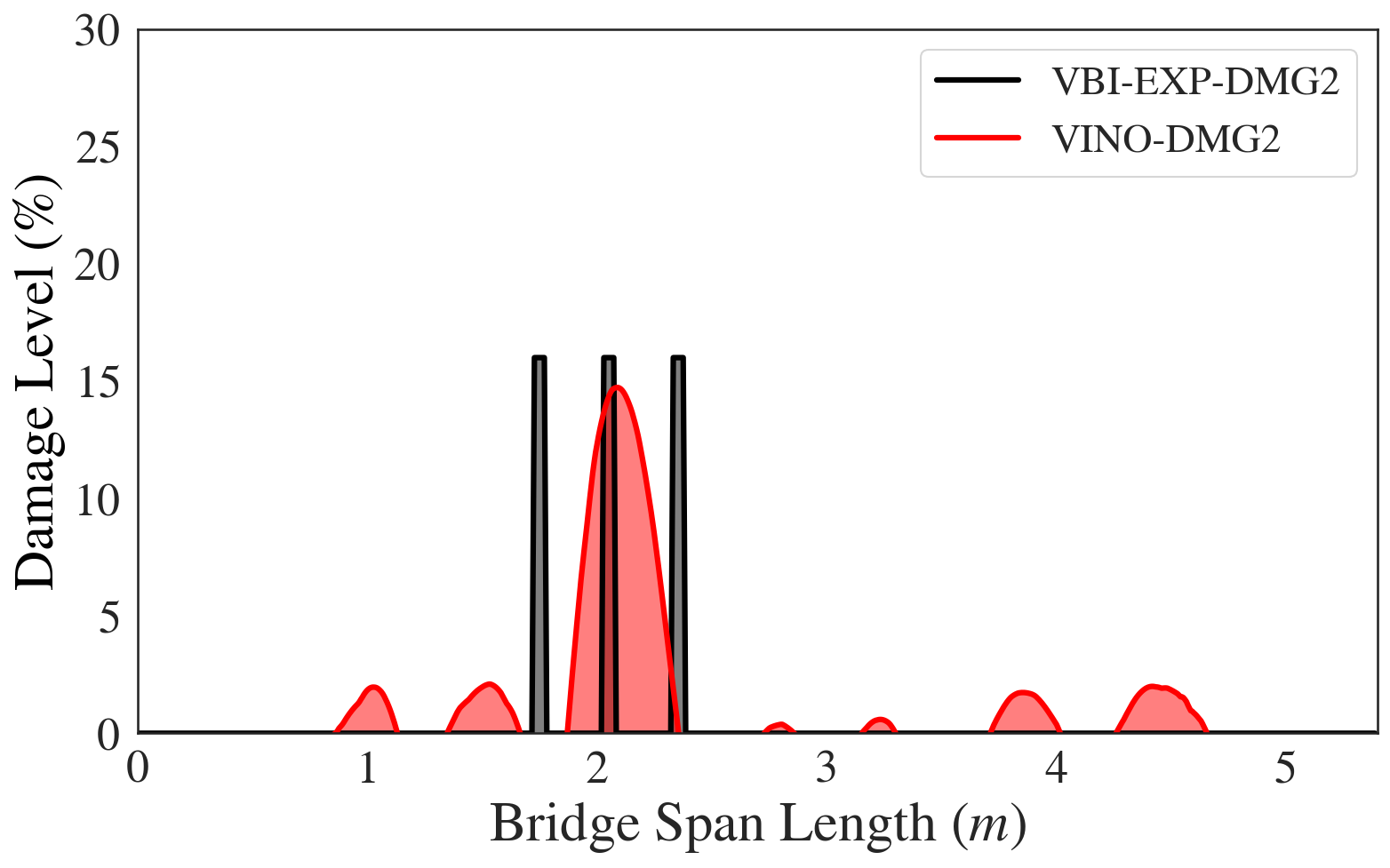}}
  \qquad
  \subfloat[Displacement responses of DMG3 scenario\label{fig:expinvdef-g}]{\includegraphics[width=0.35\linewidth]{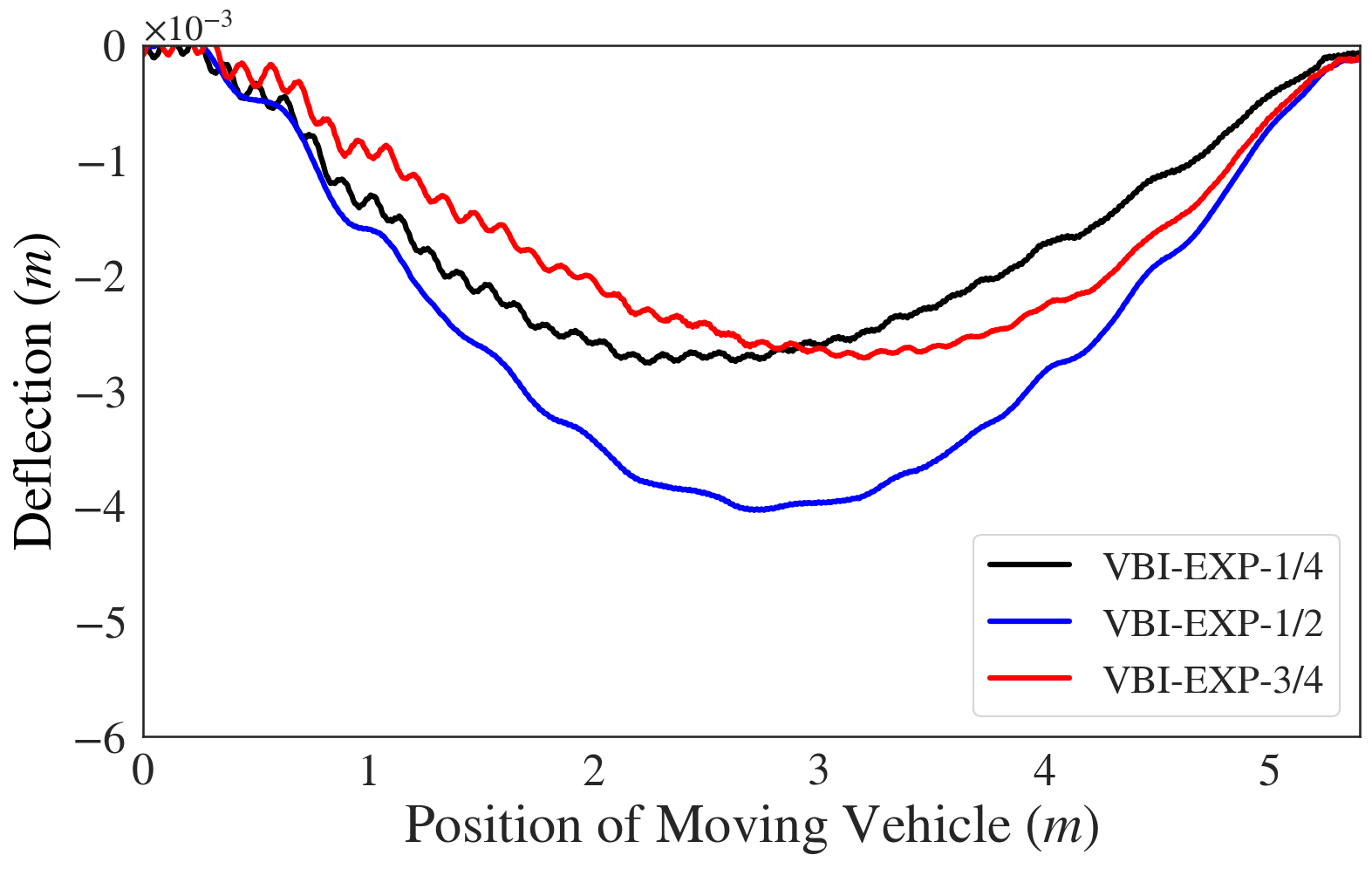}}
  \hspace{24pt}
  \subfloat[Predicted damage field of DMG3 scenario\label{fig:expinvdef-h}]{\includegraphics[width=0.35\linewidth]{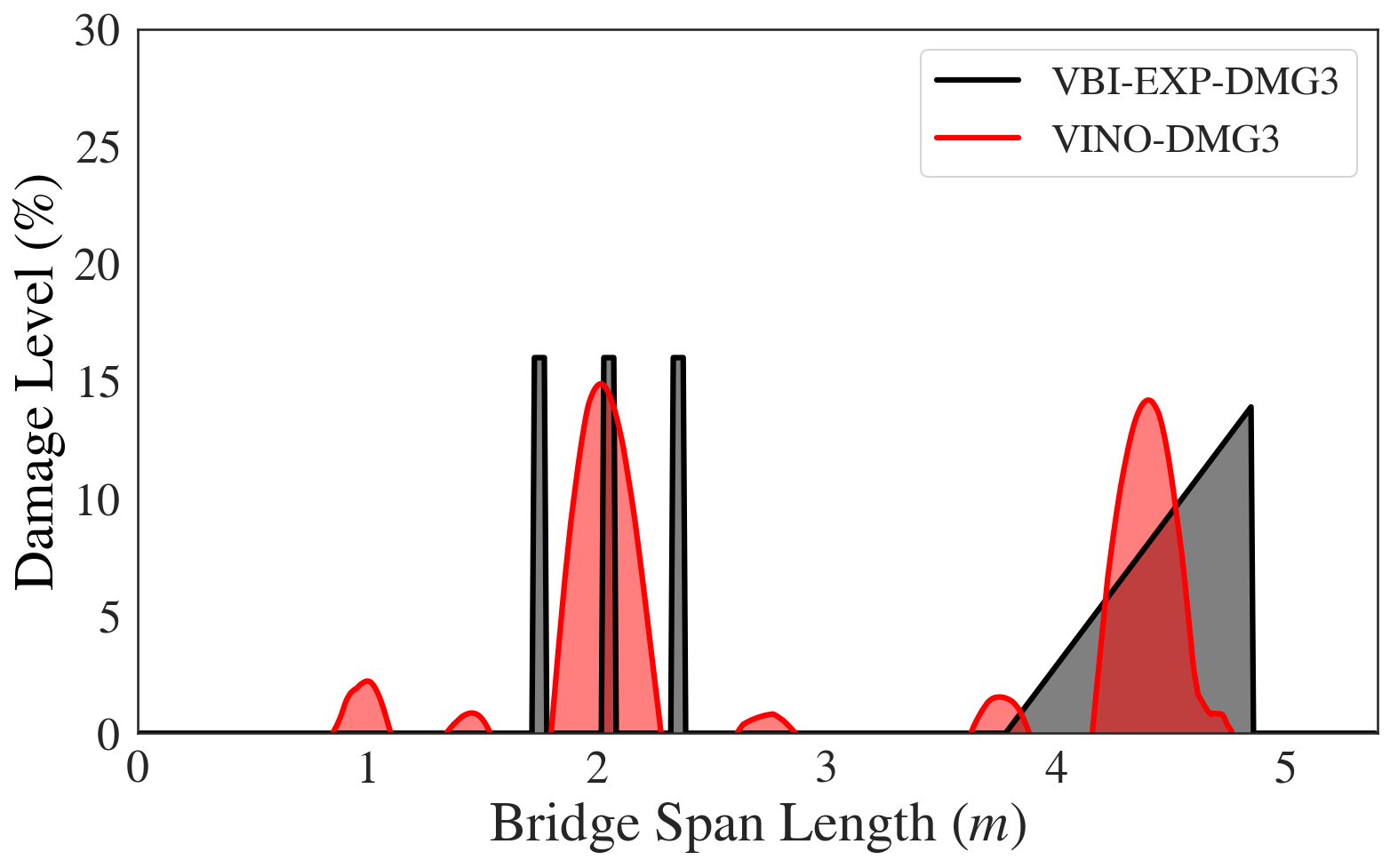}}
  \qquad
  \caption{Experimental Validation of fine-tuned inverse VINO for Structural Health Monitoring. These are comparison of predicted damages fields for INT, DMG1, DMG2, and DMG3 scenarios from the displacement fields}
  \label{fig:experimental-inverse-def}   
\end{figure*}

Figure \ref{fig:experimental-inverse-def} shows the experimental deflection curve measured by displacement sensors and predicted damage distribution of VINO compared to theoretical damage distribution. Figure \ref{fig:experimental-inverse-def} clearly demonstrated that VINO is capable of predicting all the expected damage scenarios on the laboratory bridge. In the training set, the damaged responses did not appear after fine-tuning. In the test set, the model understood damage 1 and 2 in the DMG1 and DMG2 scenarios. VINO predicts the accurate combination of damage 1 and 2 in damage 3 of the DMG3 scenario. The small damages in Figures \ref{fig:expinvdef-b}, \ref{fig:expinvdef-d}, \ref{fig:expinvdef-f}, and \ref{fig:expinvdef-h} are considered the errors at the current stage as they are considerably small (less than 5\%) compared to the expected damage (over 15\%). In addition, as observed in the figures, the spatial resolution of the damage field of VINO cannot precisely obtain the real damage field, which may be attributed to the limited number of sensors in this study. This could be the effect of both the sensors and the limited size of the training set in the VBI-FE dataset, where the damage distribution generated was very smooth, and this particular shape of damage as a piecewise function was not observed. However, the results can determine the excellent performance of VINO.

\begin{figure*}[ht!] 
\centering
  \subfloat[Acceleration responses of INT scenario\label{fig:expinvacc-a}]{\includegraphics[width=0.35\linewidth]{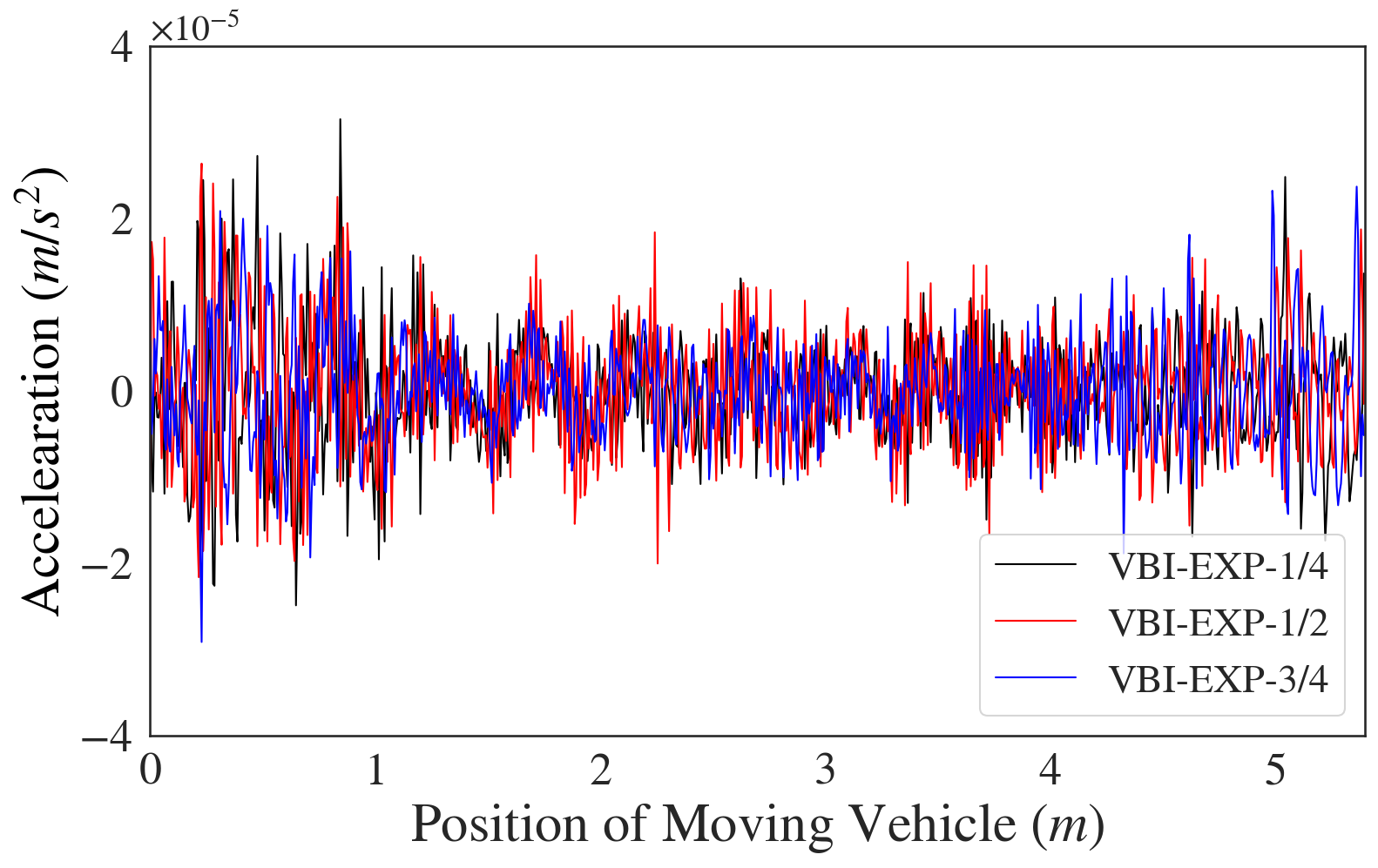}}
  \hspace{24pt}
  \subfloat[Predicted damage field of INT scenario\label{fig:expinvacc-b}]{\includegraphics[width=0.35\linewidth]{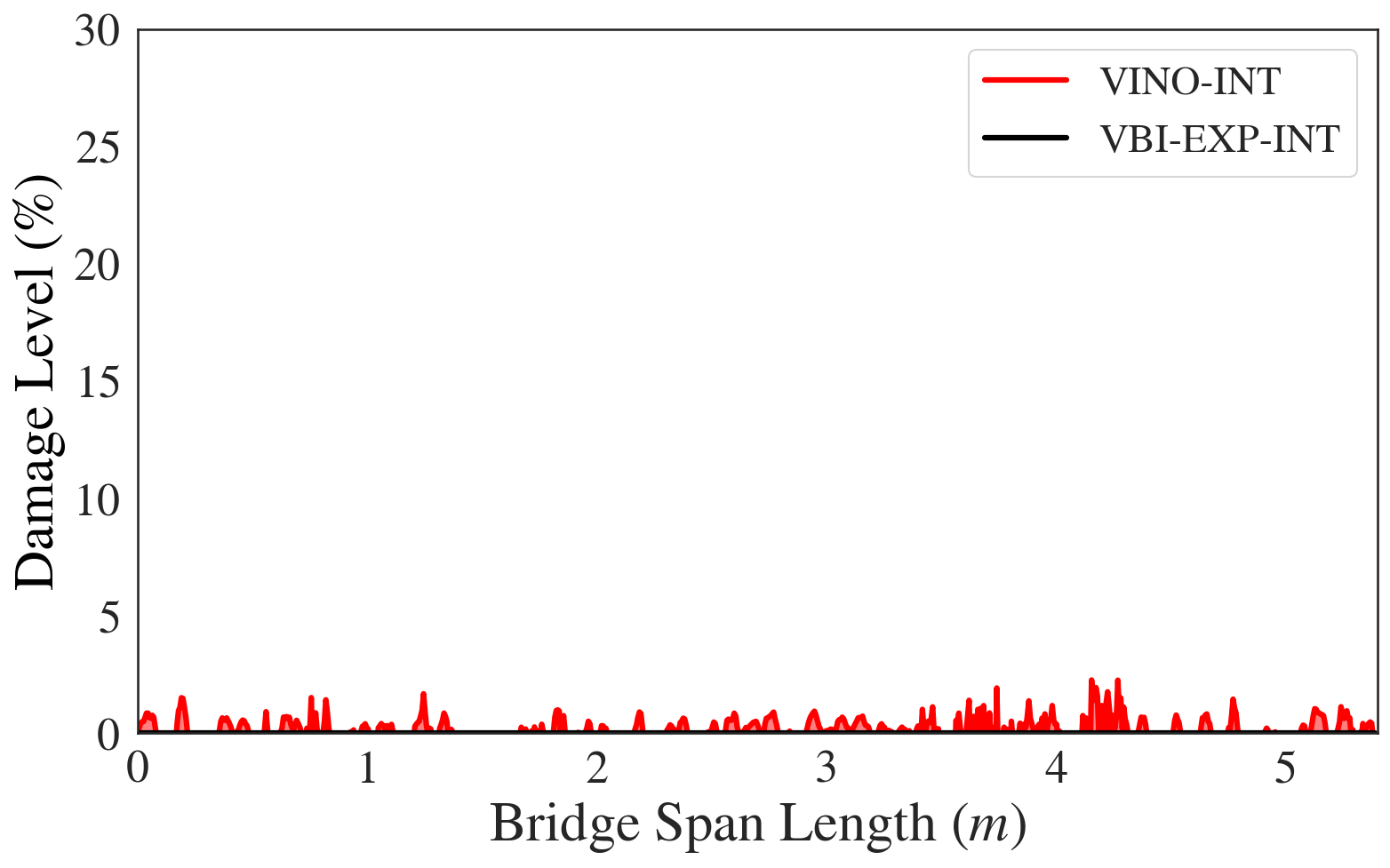}}
  \qquad
  \subfloat[Acceleration responses of DMG1 scenario\label{fig:expinvacc-c}]{\includegraphics[width=0.35\linewidth]{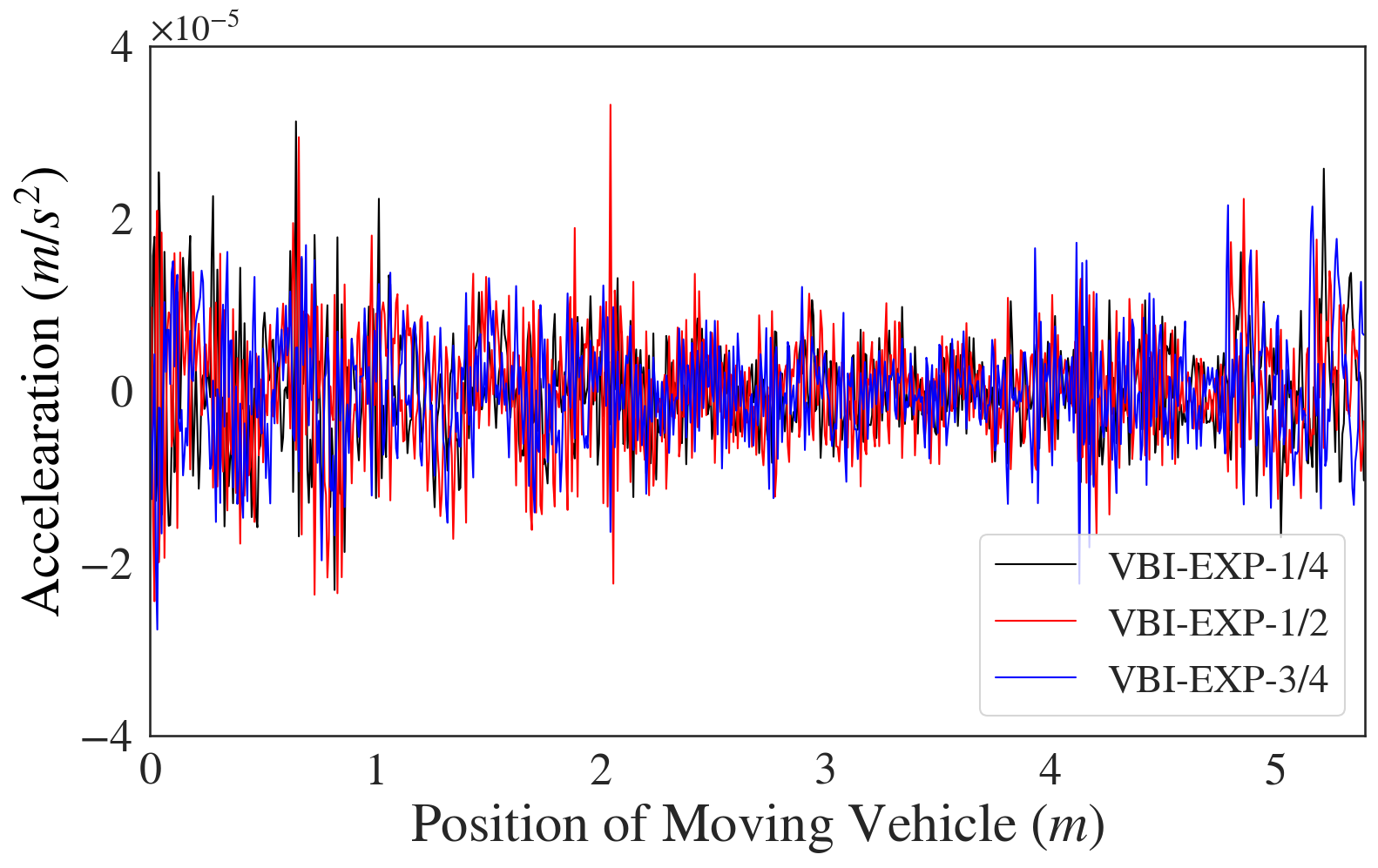}}
  \hspace{24pt}
  \subfloat[Predicted damage field of DMG1 scenario\label{fig:expinvacc-d}]{\includegraphics[width=0.35\linewidth]{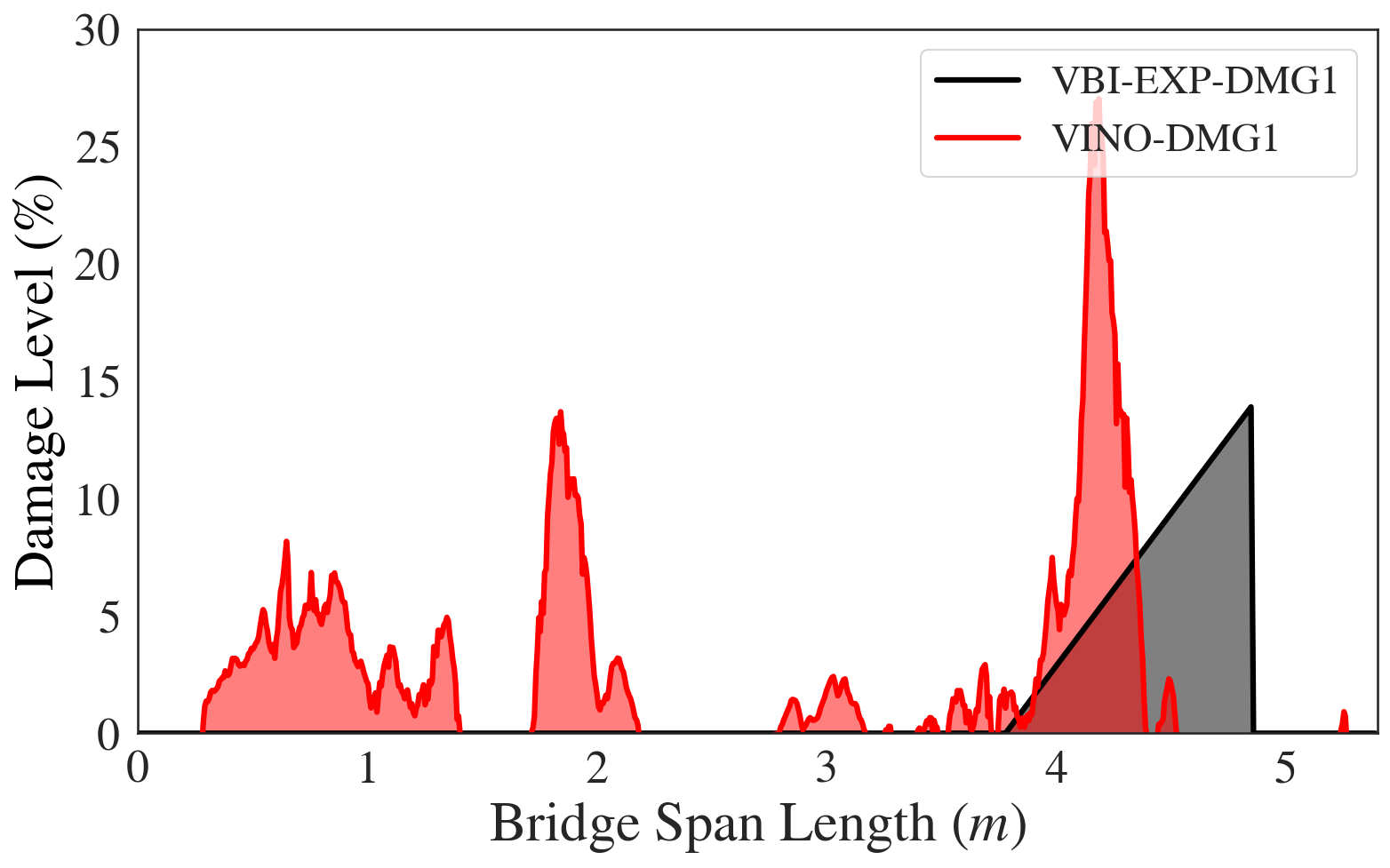}}
  \qquad
  \subfloat[Acceleration responses of DMG2 scenario\label{fig:expinvacc-e}]{\includegraphics[width=0.35\linewidth]{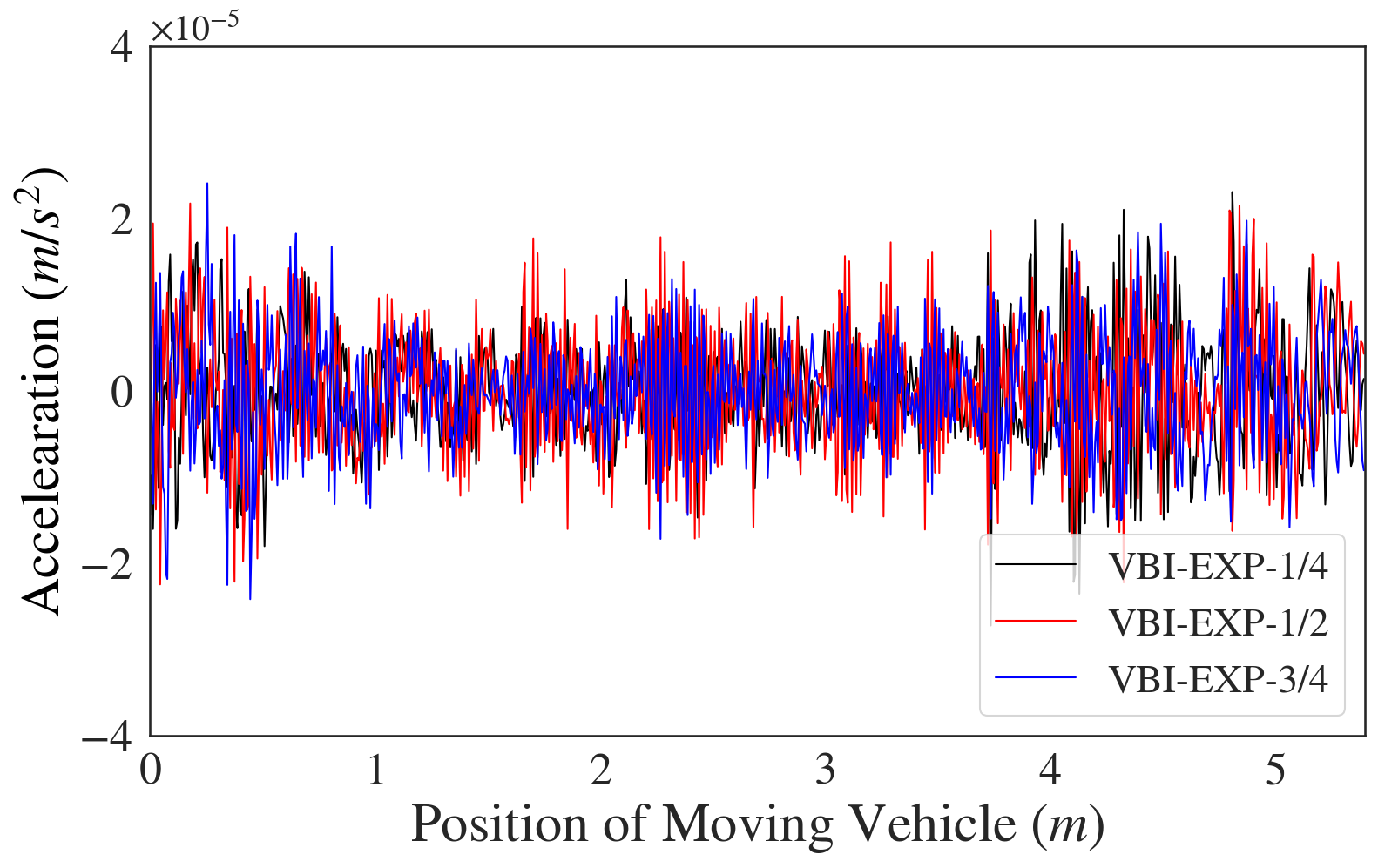}}
  \hspace{24pt}
  \subfloat[Predicted damage field of DMG2 scenario\label{fig:expinvacc-f}]{\includegraphics[width=0.35\linewidth]{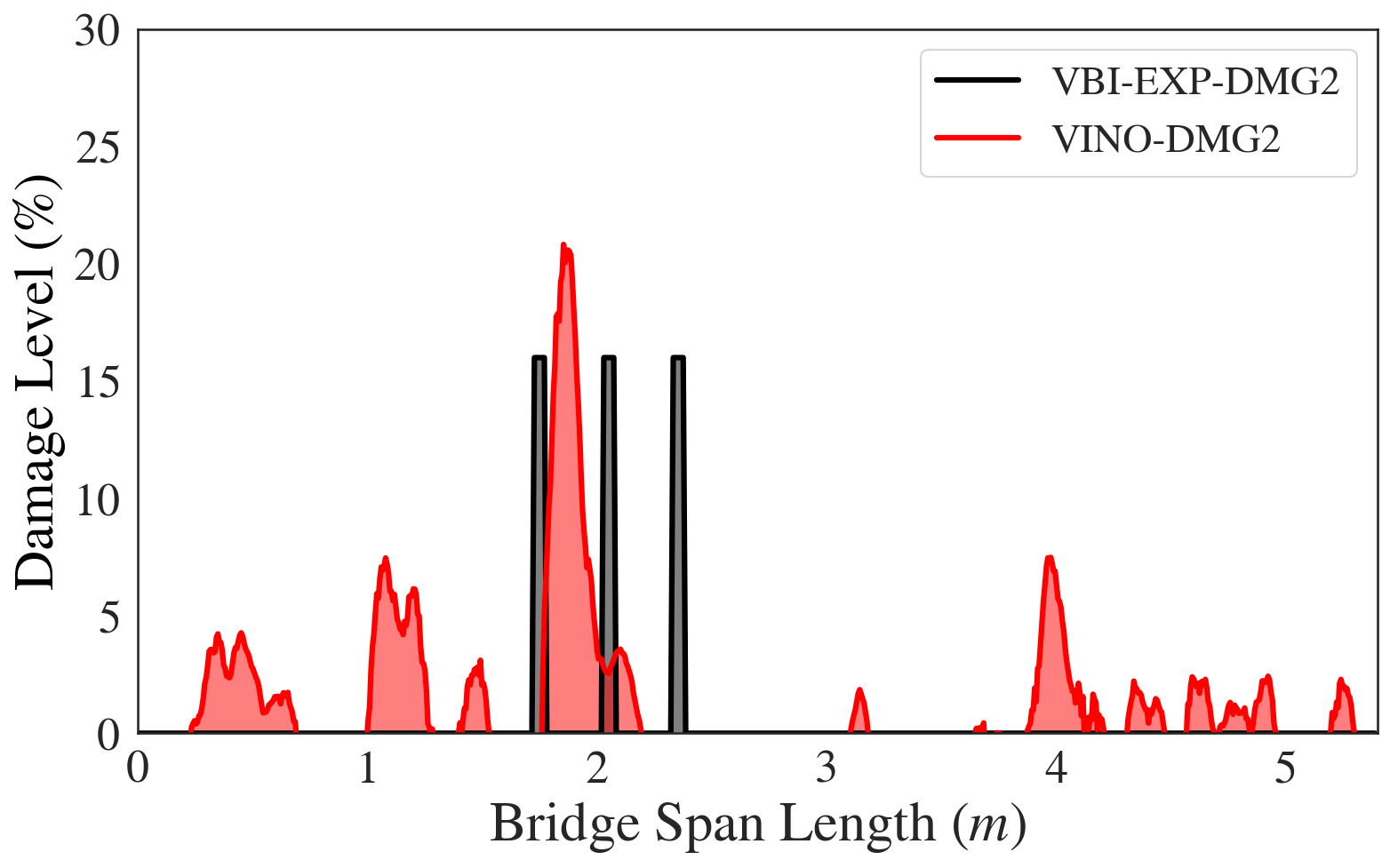}}
  \qquad
  \subfloat[Acceleration responses of DMG3 scenario\label{fig:expinvacc-g}]{\includegraphics[width=0.35\linewidth]{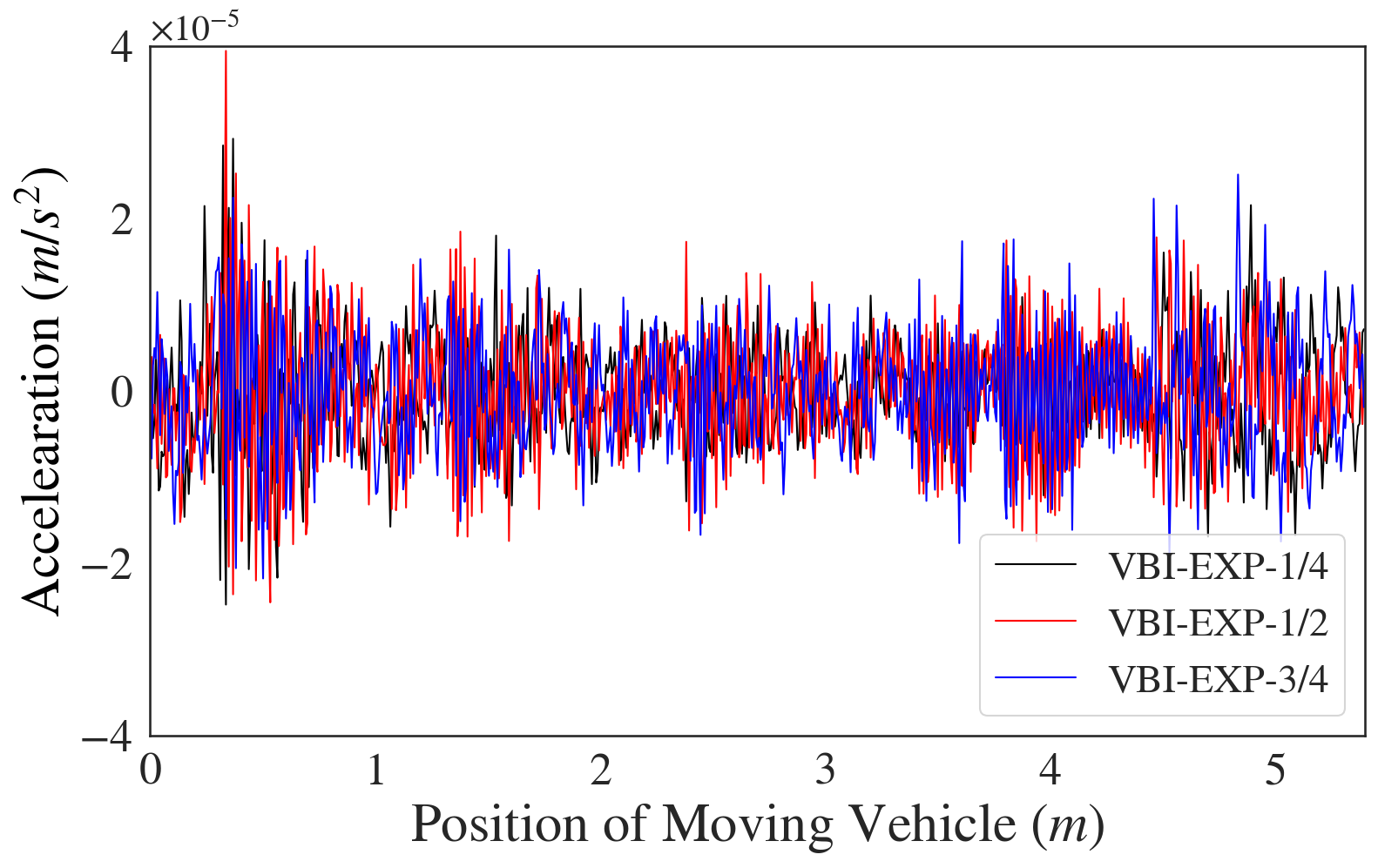}}
  \hspace{24pt}
  \subfloat[Predicted damage field of DMG3 scenario\label{fig:expinvacc-h}]{\includegraphics[width=0.35\linewidth]{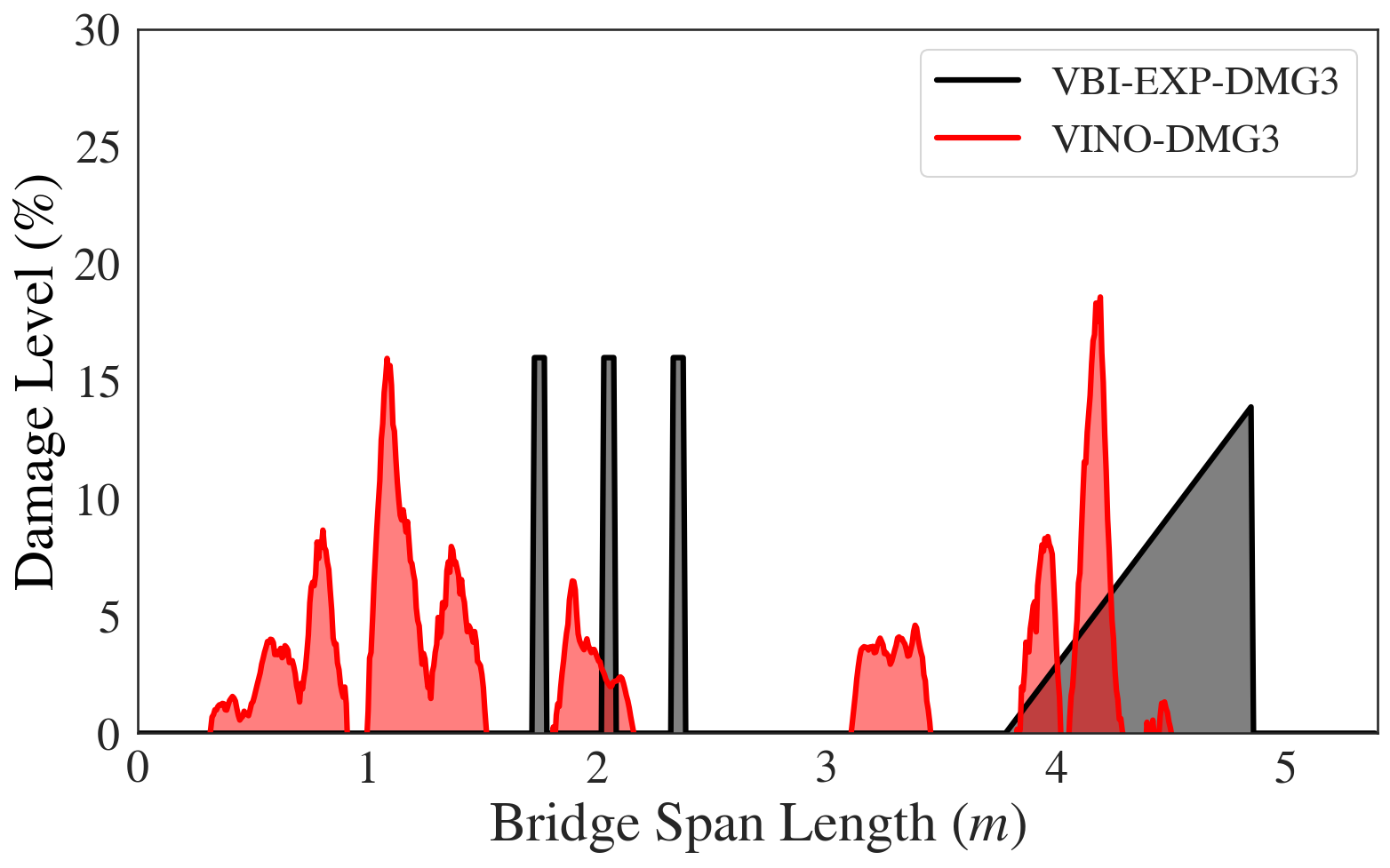}}
  \qquad
  \caption{Experimental Validation of fine-tuned inverse VINO for Structural Health Monitoring. These are the comparison of predicted damages fields for INT, DMG1, DMG2, and DMG3 scenarios from the acceleration fields.}
  \label{fig:experimental-inverse-acc} 
\end{figure*}

For the acceleration responses, the results of the investigation are shown in Figure \ref{fig:experimental-inverse-acc}. The accelerations in the time domain include high-frequency signals. Compared to the VINO trained by displacement data, the fine-tuned VINO achieved prediction of the damage field with higher errors as shown in Figures \ref{fig:expinvacc-b}, \ref{fig:expinvacc-d}, \ref{fig:expinvacc-f}, and \ref{fig:expinvacc-h}. It was observed that the VINO model for acceleration was able to predict damage 1, damage 2, and damage 3 in the DMG1, DMG2, and DMG3 scenarios to a certain level. The predictions of DMG1 and DMG2 were better than that of DMG3. The main source of error may originate from a larger noise of the signal. This suggested conducting further studies to obtain a larger experimental dataset for fine-tuning VINO models. Compared to conventional FE model updating algorithms in BHM in the time domain or frequency domain conducted by \citet{kaiqi}, VINO achieved end-to-end damage detection without the need for model updating, which notably reduced the time in damage detection.

\section{Conclusion}\label{sec:conclusion}

This paper mainly proposed the framework for the VINO approach that can serve as a digital twin of bridge structures with the VBI effect. VINO replicates a FE model with notably reduced simulation time and practicality in one- dimensional structural damage detection problems. Hence, this work contributes to the following breakthroughs of data-driven SHM using deep learning models.

\begin{enumerate}
\item Vehicle-bridge Interaction Neural Operator (VINO) model provides end-to-end, fast, and accurate in both forward (structural simulation) problems and inverse (SHM) problems.
\item For the forward problem, the FE simulation results of bridge response can be well captured by the VINO model trained with the VBI-FE dataset with negligible error. The inference speed of VINO was more than 19 times faster than a FE simulation for the prototype bridge.
\item For the forward problem, after fine-tuning from the healthy bridge training set in the VBI-EXP dataset, fine-tuned VINO achieved better structural response prediction compared to FE simulation results.
\item For the inverse problem, the VINO model trained from the VBI-FE dataset can achieve the all-in-one damage determination, localization, and quantification model.
\item For the inverse problem, the VINO model, which was fine-tuned from the healthy bridge training set in the VBI-EXP dataset, predicted damages on the test bridge efficiently with an adequate level of accuracy. 
\end{enumerate}

While the VINO model showed remarkable results, the limitations of the model include three conditions. First, the VINO model requires establishing numerical or experimental datasets. There is a strong need to develop high-fidelity datasets, including both high-fidelity numerical simulation data and experimental data, to train state-of-the-art digital twins of engineering structures. Second, to generalize to consider more parameters, such as bridge span length, cross-section size, vehicle speed, and vehicle weight, more parameters may need to be added to VINO architecture as additional static parameters. Third, the VINO model in its current form is a data-driven approach, while more physics constraints can be further added to VINO to achieve physics-informed neural networks in the future. This may improve the performance of VINO and reduce the requirement for a big dataset.

\section*{Acknowledgments}
This study is supported by the JSPS Fellowship (P22062), and JSPS Bilateral joint research projects, Grant No. JPJSBP120217405, which is greatly appreciated.

\subsection*{Author contributions}

Chawit Kaewnuratchadasorn and Jiaji Wang conceived of the presented ideas, performed computations, and carried out the experiments. Jiaji Wang encouraged Chawit Kaewnuratchadasorn to investigate the Fourier Neural Operator on the vehicle-bridge Interaction simulation data. Chul-Woo Kim supervised the finding of this work, provided critical feedback, and helped shape the research. All authors discussed the results and contributed to the final manuscript.

\subsection*{Financial disclosure}

None reported.

\subsection*{Conflict of interest}

The authors declare no potential conflict of interests.

\bibliography{reference}%

\section*{Author Biography}

\begin{biography}{\includegraphics[width=70pt,height=70pt]{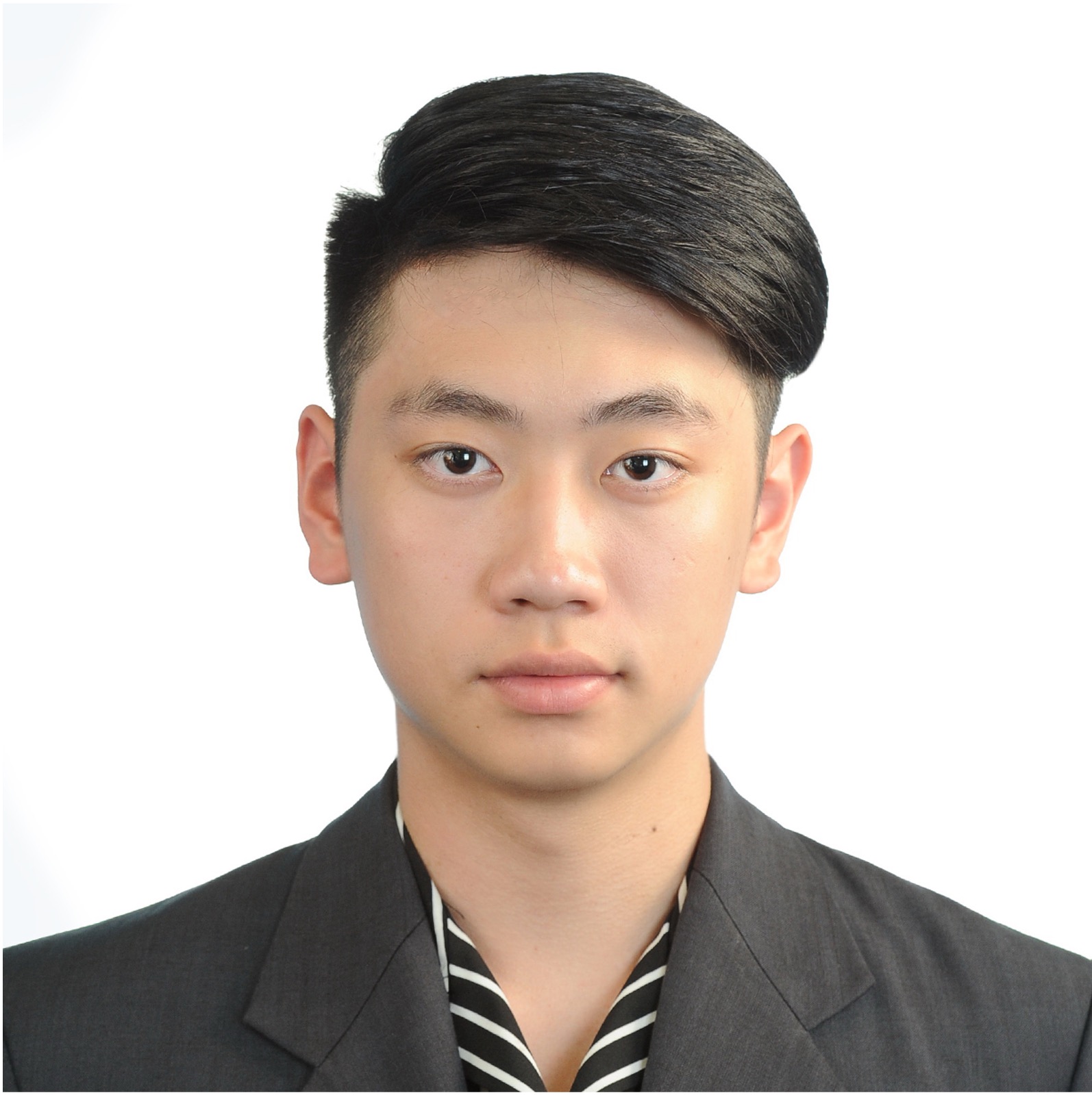}}{\textbf{Chawit Kaewnuratchadasorn} received his bachelor's degree in civil engineering from Kyoto University. He is working towards structural health monitoring at the Infrastructure Innovation Engineering Laboratory. His current research interests include machine-learning integration for anomaly detection and vision-based structural monitoring.}
\end{biography}

\begin{biography}{\includegraphics[width=70pt,height=70pt]{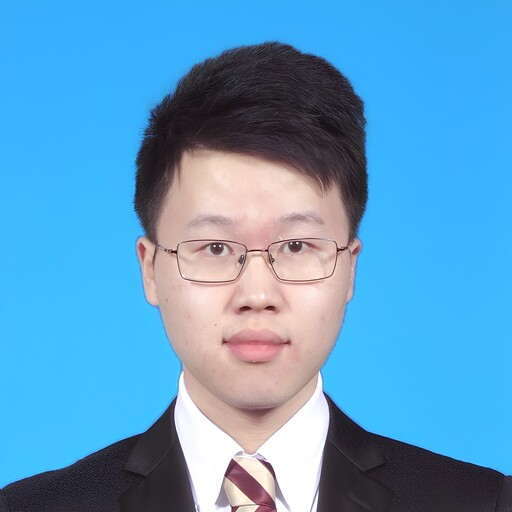}}{\textbf{Jiaji Wang} joined the Department of Civil Engineering at the University of Hong Kong as Assistant Professor in Jan 2023. He obtained his Ph.D. from Tsinghua University and served as JSPS postdoctoral fellow at Kyoto University. Enthusiastic about applying high-fidelity computational methods in structural analysis and design, Dr. Wang’s researches are mainly focused on: High fidelity constitutive models for reinforced concrete and composite structures and Physics-informed machine learning for structural engineering.}
\end{biography}

\begin{biography}{\includegraphics[width=70pt,height=70pt]{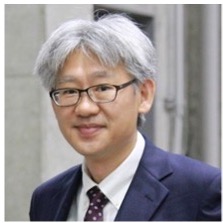}}{\textbf{Chul-Woo Kim} is the chair professor in Infrastructure Innovation Engineering Laboratory, Civil and Earth Resources Engineering at Kyoto University. He has been a professor at Kyoto University since 2009. He received his Doctor of Engineering degree from Kobe University in 2003. Prior to his appointment at Kyoto University, he worked as chief engineer in a consulting company in Japan. During that time, he was involved in cable vibration monitoring for the cable tensioning of the Stone Cutter bridge in Hong Kong. His research interests are oriented toward vehicle-bridge interactive systems to SHM of bridges, sensing for civil infrastructure condition assessment, data-driven infrastructure management, environmental vibrations propagated by bridge vibrations, structural reliability, and performance-based design, seismic performance of viaducts under traffic loadings, surrogate model in FE model update, and information fusion.}
\end{biography}

\end{document}